\documentclass{IEEEtran}
\usepackage{paralist}
\usepackage{url}
\usepackage{graphicx}
\usepackage{dblfloatfix} 
\usepackage{footmisc}
\usepackage{amssymb}
\usepackage{floatrow}
\usepackage{cite}
\usepackage{bm}
\usepackage{caption}
\usepackage{color} 
\usepackage{multirow}
\usepackage{booktabs}
\usepackage{amsmath}
\usepackage[labelformat=simple]{subcaption}
\usepackage{array}
\newcolumntype{L}[1]{>{\raggedright\let\newline\\\arraybackslash\hspace{0pt}}m{#1}}
\newcolumntype{C}[1]{>{\centering\let\newline\\\arraybackslash\hspace{0pt}}m{#1}}
\newcolumntype{R}[1]{>{\raggedleft\let\newline\\\arraybackslash\hspace{0pt}}m{#1}}
\DeclareCaptionLabelSeparator{periodspace}{.\quad}

\begin{document}

\title{Hierarchical Representation Learning for Kinship Verification}

\author{Naman~Kohli,~\IEEEmembership{Student~Member,~IEEE,}
        Mayank~Vatsa,~\IEEEmembership{Senior~Member,~IEEE,}
        Richa~Singh,~\IEEEmembership{Senior~Member,~IEEE,}
		Afzel~Noore,~\IEEEmembership{Senior~Member,~IEEE,}% <-this % stops a space
		\ and~Angshul~Majumdar,~\IEEEmembership{Senior~Member,~IEEE}

\thanks{N. Kohli and A. Noore are with the Lane Department of Computer Science
and Electrical Engineering, West Virginia University, Morgantown, WV 26506
USA e-mail: (nakohli@mix.wvu.edu; afzel.noore@mail.wvu.edu).}% <-this % stops a space
\thanks{M. Vatsa, R. Singh and A. Majumdar are with the Indraprastha Institute of Information
Technology Delhi, New Delhi 110078, India (e-mail: mayank@iiitd.ac.in;
rsingh@iiitd.ac.in; angshul@iiitd.ac.in).}% <-this % stops a space
}

\markboth{IEEE Transactions on Image Processing}%
{Kohli \MakeLowercase{\textit{et al.}}: Hierarchical Representation Learning for Kinship Verification}

\maketitle

\begin{abstract}
Kinship verification has a number of applications such as organizing large collections of images and recognizing resemblances among humans. In this research, first, a human study is conducted to understand the capabilities of human mind and to identify the discriminatory areas of a face that facilitate kinship-cues. The visual stimuli presented to the participants determines their ability to recognize kin relationship using the whole face as well as specific facial regions. The effect of participant gender and age and kin-relation pair of the stimulus is analyzed using quantitative measures such as accuracy, discriminability index $d'$, and perceptual information entropy. Utilizing the information obtained from the human study, a hierarchical Kinship Verification via Representation Learning (KVRL) framework is utilized to learn the representation of different face regions in an unsupervised manner. We propose a novel approach for feature representation termed as filtered contractive deep belief networks (\textit{fc}DBN). The proposed feature representation encodes relational information present in images using filters and contractive regularization penalty. A compact representation of facial images of kin is extracted as an output from the learned model and a multi-layer neural network is utilized to verify the kin accurately. A new WVU Kinship Database is created which consists of multiple images per subject to facilitate kinship verification. The results show that the proposed deep learning framework (KVRL-\textit{fc}DBN) yields state-of-the-art kinship verification accuracy on the WVU Kinship database and on four existing benchmark datasets. Further, kinship information is used as a soft biometric modality to boost the performance of face verification via product of likelihood ratio and support vector machine based approaches. Using the proposed KVRL-\textit{fc}DBN framework, an improvement of over 20\% is observed in the performance of face verification. 
\end{abstract}

\begin{IEEEkeywords}
Kinship Verification, Face Verification, Deep Belief Networks, Soft Biometrics
\end{IEEEkeywords}

\IEEEpeerreviewmaketitle
%
%\setcounter{tocdepth}{4}
%\tableofcontents 

%\newpage
\graphicspath{{./Images/}}

\section{Introduction}

\IEEEPARstart{K}{inship} refers to sharing of selected characteristics among organisms through nature. Kinship verification is the task of judging if two individuals are kin or not and has been widely studied in the field of psychology and neuroscience. Hogben \cite{lancelot} called the similarities in facial structure of humans as familial traits. Face resemblance is thought to be one of the most common physical cues for kinship \cite{debruinesocial2008}. The hypothesis that similarity among faces could be a cue for kinship was first formulated by Daly and Wilson \cite{Daly1982}. Since then, facial similarity/resemblance has been used to judge kinship recognition in a number of research experiments \cite{Christenfeld1995, Bredart1999, Bressan2002, burch2000, Platek2004, KaminskyK2009, McLain2000, Oda2005}. Maloney and Martello \cite{Maloney2006} have examined the relation between \textit{similarity} and \textit{kinship detection} among siblings and concluded that observers do look for similarity in judging kinship among children. Martello and Maloney \cite{Martello2006} have further shown that in kinship recognition, the upper portion of a face has more discriminating power as compared to the lower half. In a different study, to determine the effect of lateralization on allocentric kin recognition, they have suggested that the right half of the face is equal to the left half portion of the face for the purpose of kinship recognition\cite{MartelloM2010}.

\begin{table*}[!t]
\centering
\small
\begingroup
\renewcommand*{\arraystretch}{1.2}
\begin{tabular}{|c|L{3cm}|L{4.5cm}|l|C{4.5em}|C{5em}|}
\hline
\textbf{Year} & \textbf{Authors} & \textbf{Algorithm} & \textbf{Database} & \textbf{Accuracy (\%) } & \textbf{Outside Training}\\
\hline
\hline
2010                                        & Fang et al. \cite{Fang2010}                       & Pictorial structure model                                                  & Cornell KinFace           & 70.67  & \multirow{28}{*}{No} \\ \cline{1-5}
\multirow{3}{*}{2011}                       & Siyu et al. \cite{Siyu2011}                        & Transfer learning                                                          & UB Kin Database           & 60.00 &  \\ \cline{2-5} 
                                            & Shao et al.  \cite{Ming_CVPR11_Genealogical}                      & Transfer subspace learning                                                 & UB Kin Database           & 69.67 &  \\ \cline{2-5} 
                                            & Zhou et al  \cite{Zhou2011}                       & Spatial pyramid learning based kinship                                     & Private Database& 67.75 &  \\ \cline{1-5}
\multirow{5}{*}{2012}                       & Xia et al. \cite{Buffalo_TMM_Kinship}                        & Attributes LIFT learning                                                   & UB Kin Database           & 82.50 &   \\ \cline{2-5} 
                                            & \multirow{2}{4cm}{Kohli et al.\cite{Kohli2012}}      & \multirow{2}{4cm}{Self similarity representation of weber faces}             & UB Kin Database           & 69.67 &  \\ \cline{4-5} 
                                            &                                    &                                                                            & IIITD Kinship Database    & 75.20 &  \\ \cline{2-5} 
                                            & Guo et al \cite{Guo2012}                         & Product of likelihood ratio on salient features                            & Private Database& 75.00 &  \\ \cline{2-5} 
                                            & Zhou et al.  \cite{Zhou2012}                      & Gabor based gradient oriented pyramid                                      & Private Database& 69.75 &  \\ \cline{1-5}
2013                                        & Dibeklioglu et al.  \cite{Dibeklioglu}               & Spatio temporal features                                                   & UvA-NEMO Smile    & 67.11 &   \\ \cline{1-5}
\multirow{12}{*}{2014}                       & \multirow{2}{*}{Lu et al. \cite{lu}}         & \multirow{2}{4cm}{Multiview neighborhood repulsed metric learning}           & KinFace-I                 & 69.90 &  \\ \cline{4-5} 
                                            &                                    &                                                                            & KinFace-II                & 76.50 &  \\ \cline{2-5} 
                                            & \multirow{4}{*}{Yan et al.\cite{Yan}}        & \multirow{4}{4cm}{Discriminative multimetric learning}                       & Cornell KinFace           & 73.50* &  \\ \cline{4-5} 
                                            &                                    &       & UB Kin Database           & 74.50 &  \\ \cline{4-5} 
                                            &                                    &        & KinFace-I                 & 72.00* & \\ \cline{4-5} 
                                            &                                    &           & KinFace-II                & 78.00* & \\ \cline{2-5} 
                                            & \multirow{2}{*}{Dehghan et al. \cite{dehghanOVS14}}    & \multirow{2}{4cm}{Discrimination via gated autoencoders}                     & KinFace-I                 & 74.50 &   \\ \cline{4-5} 
                                            &                                    &              & KinFace-II                & 82.20 &  \\ \cline{2-5}
\multicolumn{1}{|l|}{} & \multirow{4}{*}{Yan et al. \cite{Yan2014} }        & \multirow{4}{4cm}{Prototype discriminative  feature learning}                & Cornell KinFace           & 71.90 & \\ \cline{4-5} 
\multicolumn{1}{|l|}{}                      &                                    &            & UB Kin Database           & 67.30 &  \\ \cline{4-5} 
\multicolumn{1}{|l|}{}                      &                                    &           & KinFace-I                 & 70.10 &  \\ \cline{4-5} 
\multicolumn{1}{|l|}{}                      &                                    &           & KinFace-II                & 77.00 &   \\ \cline{1-5} 

\multicolumn{1}{|l|}{\multirow{6}{*}{2015}}                      & \multirow{2}{*}{Liu et al. \cite{fisher_kinship}} & \multirow{2}{4cm}{Inheritable Fisher Vector Feature based kinship} & KinFace-I           & 73.45  &  \\ \cline{4-5} 
\multicolumn{1}{|l|}{}                      &                                    &            & KinFace-II           & 81.60 &  \\ \cline{2-5}
\multicolumn{1}{|l|}{}                      &    \multirow{2}{*}{Alirezazadeh et al. \cite{genetic_kinship}}               &  \multirow{2}{4cm}   {Genetic Algorithm for feature selection for kinship}       & KinFace-I           & 81.30 &  \\ \cline{4-5} 
\multicolumn{1}{|l|}{}                      &                                    &            & KinFace-II           & 86.15 &  \\ \cline{2-5} 
\multicolumn{1}{|l|}{}                      &    \multirow{2}{*}{Zhou et al. \cite{Zhou2015}}               &  \multirow{2}{4cm}   {Ensemble similarity learning}       & KinFace-I           & 78.60 & \\ \cline{4-5} 
\multicolumn{1}{|l|}{}                      &                                    &            & KinFace-II           & 75.70 &  \\ \cline{4-5} \hline

\multicolumn{1}{|l|}{\multirow{5}{*}{2016}}                      & \multirow{5}{*}{\textbf{Proposed}} & \multirow{5}{4cm}{\textbf{Kinship verification via representation learning (KVRL-\textit{fc}DBN)}} & Cornell KinFace           & \textbf{89.50}  & \multirow{5}{*}{Yes} \\ \cline{4-5} 
\multicolumn{1}{|l|}{}                      &                                    &            & UB Kin Database           & \textbf{91.80} &  \\ \cline{4-5} 
\multicolumn{1}{|l|}{}                      &                                    &              & KinFace-I                 & \textbf{96.10}  &  \\ \cline{4-5} 
\multicolumn{1}{|l|}{}                      &                                    &               & KinFace-II                &\textbf{96.20}  &  \\ \cline{4-5} 
\multicolumn{1}{|l|}{}                      &                                    &               & WVU Kinship Database      & \textbf{90.80}  &  \\ \hline
\end{tabular} 
\endgroup
%\captionsetup{justification=centering}
\caption{Review of kinship verification algorithms. Outside Training column represents if an external face database was required for training the algorithm.
The symbol * represents value taken from ROC curve}
\label{tab:review_table}

\end{table*}

\begin{figure}[!b]
	\centering
		\includegraphics[width=1\linewidth]{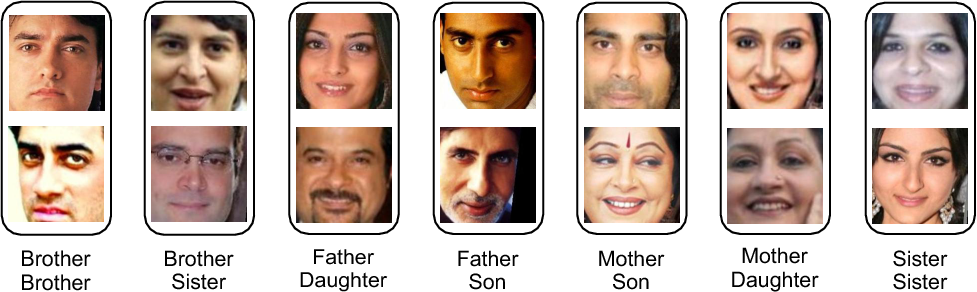}
		
	\captionsetup{justification=centering}
	\caption{Examples of kin-relations considered in this research.}
	\label{fig:ExampleOfKins}
\end{figure}

Some examples of kin-relations are shown in Fig. \ref{fig:ExampleOfKins}. Kinship verification has several applications such as:

\begin{enumerate}
	\item organizing image collections and resolving identities in photo albums,
	\item searching for relatives in public databases,
	\item boosting automatic face verification capabilities,
	\item automatically tagging large number of images available online, and
	\item finding out kin of a victim or suspect by law enforcement agencies.
\end{enumerate}

Kinship verification has also gained interest in the computer vision and machine learning communities. The first dataset containing kin pairs was collected by Fang et al. \cite{Fang2010}. For performing kinship verification, the authors proposed an algorithm for facial feature extraction and forward selection methodology. Since then, the algorithms for verifying kin have increased in complexity and Table \ref{tab:review_table} provides a review of algorithms recently published in this area along with the databases used. The problem of kinship verification is particularly challenging because of the large intra-class variations among different kin pairs. At the same time, look-alikes decrease the inter-class variation among the facial images of kin. While existing algorithms have achieved reasonable accuracies, there is a scope of further improving the performance. For instance, deep learning algorithms can be utilized; however, they typically require a large training database which existing kinship databases lack. Moreover, kinship cues can be visualized as the \textit{soft} information that can be utilized to boost the performance of face verification algorithms.

\subsection{Research Contributions}
Inspired by face recognition literature, where researchers have tried to understand how humans perform face recognition, we have performed a similar study to understand the ability of humans in identifying kin. Using the cues from human study, this research presents a deep learning based kinship verification framework that relies on learning face representations. A new approach using the proposed filtered contractive deep belief networks (\textbf{\textit{fc}DBN}) is presented where the formulation of RBMs is extended through a filtering approach and a contractive penalty. The idea of this approach stems from the fact that facial images have an inherent structure which can be emphasized using filters. By simultaneously learning filters and weights, an invariant representation of the faces is learned which is utilized in kinship verification. Using contractive penalty, we learn robust features that are invariant to local variations in the images. The proposed approach shows state-of-the-art results on multiple datasets used in kinship verification research. 
\\Humans utilize contextual information in identifying faces such as establishing the identity of a person through kinship cues. Inspired by this phenomenon, our research models kinship as \textit{soft} information which can help in improving the performance of a strong biometric matcher. Therefore, we also present an approach that incorporates kinship as a soft biometric information for boosting the results of face verification. A new database consisting of multiple images of kin has also been created to help in evaluating the performance of the proposed kinship verification algorithm.

\section{Evaluating Human Performance for Kinship Verification}

In face recognition literature, several studies have been performed to understand the recognition capabilities of human mind. Inspired by these studies, in this research, a human study is conducted to understand the ability of humans in identifying kin. The goal of this study is to (a) understand the underlying cues that humans use to identify kin, and (b) integrate these findings in automatic deep learning algorithm to achieve better kinship verification accuracy. Lu et al. \cite{lu} have performed a similar human study based on kinship verification. They have focused specifically on the overall kinship verification accuracy and concluded that using contextual information such as hair and background improves kinship verification.

\subsection{Experimental Protocol and Databases Used}

Amazon MTurk is an online platform specifically designed for aiding research by organizing surveys and collecting results in a comprehensive manner. MTurk allows crowdsourcing and enables researchers to include participants across diverse demographics. It has been shown to provide reliable data as compared to data provided by the traditional means of survey collection and offers a rich pool of participants \cite{Buhrmester2011}. It allows the creation of Human Intelligence Tasks (HITs) for surveys, studies, and experiments which are in turn completed by participants. The participants receive a reward for completing a HIT if their results are approved by the requester. In this study conducted on Amazon MTurk, a total of 479 volunteers (200 male and 279 female) participated. Among all the participants, 366 were Indians (Mean Age (M) = 33.45 years, Standard Deviation in Age (SD) = 11.67 years), 81 were Caucasians (M = 35.39 years, SD = 10.74 years), 29 were Asians (non-Indians) (M = 28.13 years, SD = 6.93 years), and 3 were African-Americans (M = 30.33 years, SD = 8.17 years).

The images used in this study are collected from three databases: Vadana \cite{SomanathRK11,Somnath2012}, Kinship Verification database \cite{Fang2010}, and UB Kin database \cite{Buffalo_TMM_Kinship,Siyu2011,Ming_CVPR11_Genealogical}. The database consists of 150 kin pairs and 150 non-kin pairs with 39 Sister-Sister (SS) combinations, 36 Brother-Sister (BS) combinations, 35 Brother-Brother (BB) combinations, 50 Father-Son (FS) combinations, 40 Father-Daughter (FD) combinations, 41 Mother-Daughter (MD) combinations, and 59 Mother-Son (MS) combinations. Each participant is shown five pairs of images that are assigned in a random order. The participant has to answer if the subjects in the given pair of images appear to be kin to each other or not. Additionally, the participants are also asked if they have seen the subjects prior to the study. This allows us to evaluate the differences in the responses based on the familiarity with the stimuli.

Generally, the studies evaluating the human performance have used full faces. However, it is not necessary that the whole face contributes in determining kinship. Therefore, we also perform the experiments with specific facial regions. The performance of the participants is determined for the following visual stimulus: 

\begin{enumerate}
	\item full face, 
	\item T region (containing nose and eyes),
	\item not-T region (containing the face with the eye and nose regions obfuscated),
	\item lower part of facial image, and
	\item binocular region (eye strip).
\end{enumerate}

\begin{figure}[!b]
	\centering
		\includegraphics[width=0.95\linewidth]{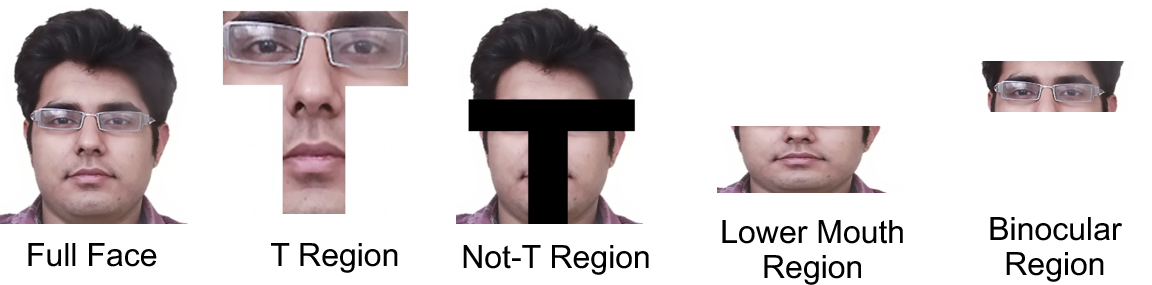}
	%\captionsetup{justification=centering}
	\caption{Sample images demonstrating seven kin-relations considered in this research.}
	\label{fig:ExampleOfFaceRegions}
\end{figure}

Fig. \ref{fig:ExampleOfFaceRegions} illustrates different facial regions extracted from faces of subjects. The binocular region is chosen to observe the effect of eyes on kinship verification. The T region represents features in the region of the face around the eyes and nose. Furthermore, to observe the effect of outer facial regions, not-T region is chosen (which does not have regions that are included in the T region). The lower facial region is included to evaluate a hypothesis stated in an earlier research study \cite{Martello2006} which claims that kinship cues are not present in this region.

\begin{table*}[!htp]
	\centering
	\small
	\begin{tabular}{|c|L{2.75cm}|C{1cm}|C{1.7cm}|C{2.5cm}|C{2cm}|C{3.25cm}|}
		\hline
		S. No. & Experiments  & $d'$       & Kin Entropy   & Non-Kin Entropy   & Total Entropy & Overall Accuracy (in \%) \\ \hline\hline
		
		\multicolumn{7}{|l|}{\textit{\textbf{Participant's Demographic - Gender}}}                                     \\ \hline
		1.  & Female                  &\textbf{0.3703}   & 0.0063        & 0.0069            & \textbf{0.0132}&    \textbf{56.00}     \\ \hline
		2.  & Male                    & 0.2982   & 0.0045        & 0.0048            & 0.0093     &  55.00  \\ \hline\hline
		\multicolumn{7}{|l|}{\textit{\textbf{Participant's Demographic - Age}}}                                        \\ \hline
		1.  & \textless30             & 0.3498   & 0.0056        & 0.0061            & 0.0117      & 55.60  \\ \hline
		2.  & 30 - 50                 & 0.3119   & 0.0050        & 0.0053            & 0.0102      & 55.51  \\ \hline
		3.  & \textgreater50          & \textbf{0.3986 }  & 0.0077        & 0.0082            & \textbf{0.0159}   &   \textbf{56.95}   \\ \hline\hline
		%\multicolumn{7}{|l|}{\textit{\textbf{Ethnicity of Participant}}}                                  \\ \hline
		%1.  & Caucasian               & \textbf{0.4948}   & 0.0132        & 0.0136            & \textbf{0.0268}  &   \textbf{59.45}    \\ \hline
		%2.  & Indian                  & 0.4266   & 0.0099        & 0.0065            & 0.0164    &  54.89   \\ \hline
		%3.  & Asian                   & 0.3914   & 0.0064        & 0.0074            & 0.0137   &    55.22  \\ \hline\hline
		\multicolumn{7}{|l|}{\textit{\textbf{Stimulus Kin Relationship}}}                                          \\ \hline
		1.  & Mother-Son                      & \textbf{0.8211}   & 0.0162        & 0.0383            & \textbf{0.0545}   &   55.39   \\ \hline
		2.  & Sister-Sister                      & 0.5505   & 0.0181        & 0.0059            & 0.0240    &  \textbf{66.23}   \\ \hline
		3.  & Father-Daughter                      & 0.3762   & 0.0065        & 0.0072            & 0.0137    &   56.01  \\ \hline
		4.  & Mother-Daughter                      & 0.3088   & 0.0046        & 0.0051            & 0.0097    &   54.74  \\ \hline
		5.  & Brother-Sister                      & 0.2482   & 0.0024        & 0.0035            & 0.0059    &   50.11  \\ \hline
		6.  & Father-Son                      & 0.2092   & 0.0021        & 0.0023            & 0.0044    &   53.48  \\ \hline
		7.  & Brother-Brother                      & 0.0560   & 0.0002        & 0.0001            & 0.0003    &   54.10  \\ \hline\hline
		\multicolumn{7}{|l|}{\textbf{\textit{Local and Global Regions of Face}}}                                            \\ \hline
		1.  & Face                    & \textbf{0.4531}   & 0.0107        & 0.0115            & \textbf{0.0221}      &  \textbf{58.36} \\ \hline
		2.  & Not-T                   & 0.4212   & 0.0084        & 0.0093            & 0.0177     &  57.02  \\ \hline
		3.  & T                       & 0.3466   & 0.0059        & 0.0064            & 0.0123     &  55.92  \\ \hline
		4.  & Chin                    & 0.2772   & 0.0037        & 0.0040            & 0.0077      &  54.57 \\ \hline
		5.  & Binocular               & 0.1656   & 0.0013        & 0.0013            & 0.0026       &  52.58  \\ \hline
	\end{tabular}
	\captionsetup{justification=centering}
	\caption{Quantitative analysis of human performance on kinship verification.}
	\label{tab:table_humanResult}
\end{table*}

\subsection{Results and Analysis}

In the human study, we analyze (a) the effect of gender and age demographics of participants on kinship verification, (b) the types of kinship relation between stimuli, and (c) the discriminative local and global face features that humans rely on to correctly verify kinship. 

Based on the responses from participants, a quantitative analysis of the data is performed using three independent measures: accuracy of correct kinship verification, discriminability or sensitivity index ($d'$), and information theory to compute the kin entropy and non-kin entropy. Discriminability or sensitivity index ($d'$) is used in signal detection theory to quantify the difference between the mean signal and noise distributions in a given stimulus as perceived by participants. %In this experiment, participants had to make a judgment regarding the discriminability of kin vs non-kin among the stimuli. The strongest response among the stimuli is considered to be the signal that the participants are looking for; while an incorrect choice is considered to be due to a noisy signal. The distance between the means of the signal and the noise are evaluated against the standard deviation of the noise distribution, to compute the discriminability index $(d')$. Higher values of $d'$ signify a better perceptible signal in that particular category.

%Another way to analyze the results is to quantify the amount of perceived information using information entropy. 

There is an inherent uncertainty in determining the relationship between stimuli. This uncertainty can be attributed to noise and higher response categories. The stimulus information entropy $H(S)$ and noise in the signal $H(S$$\mid$$r)$ are computed from the confusion matrix using Eq. \ref{eq:entropy} and \ref{eq:noise} respectively. 

%The stimulus entropy $H(S)$,  (Equation \ref{eq:entropy}), noise in the signal $H(S$$\mid$$r)$ (Equation \ref{eq:noise}) are calculated from the confusion matrix where $r$ refers to responses of participants and $S$ refers to the stimulus.  The information entropy $I(S$$\mid$$r)$ is calculated by subtracting the noise in the signal from the stimulus entropy as shown in Equation \ref{eq:infEntropy}. The information entropy is divided by $\log{2}$ to represent in bits and the results are summarized in Table \ref{tab:table_humanResult}. Larger values of the bits determine higher perceptual judgment of the participants. 

\begin{equation}
	H(S) = - \sum_{i=1}^{n} p(S_{i}) log(p(S_{i}))
	\label{eq:entropy}
\end{equation}

\begin{equation}
	H(S|r) = - \sum_{i=1}^{n} \sum_{j=1}^{n} p(S_{i},r_{j}) log( p(S_{i}|r_{j}) )
	\label{eq:noise}
\end{equation}

\begin{equation}
	I(S|r) = H(S) - H(S|r)
	\label{eq:infEntropy}
\end{equation}

\noindent Here, $r$ refers to the response of participants and $S$ refers to the stimulus. The information entropy $I(S$$\mid$$r)$ is calculated by subtracting the noise in the signal from the stimulus entropy as shown in Eq. \ref{eq:infEntropy}. The information entropy is divided by $\log{2}$ to represent in bits and larger values of the bits determine higher perceptual judgment of the participants. Higher values in accuracy or $d'$ or total entropy indicate that the signals can be more readily detected compared to other visual artifacts that do not contribute to the kinship verification.

The results are analyzed to understand the effect of four different attributes on kinship verification: gender and age of participants, relation between stimuli kin pairs, and facial regions presented in the stimuli. The results are summarized in Table \ref{tab:table_humanResult}.

\subsubsection{Effect of Participant's Gender on Kinship Verification}

In face recognition, several studies have demonstrated that women outperform men in the ability to recognize faces \cite{Rehnman_Herlitz_2007,Rehnman_2006}. In a meta-analysis study of over 140 face recognition studies, Herlitz and Loven \cite{herlitz2013} have found that females consistently outperform males in recognizing faces. This fact is also supported in \cite{susilo2013}, where females performed better than males in the face recognition task. The effect of participant's gender is analyzed to determine if there exists any difference in the skills of males and females for kinship verification. As shown in Table \ref{tab:table_humanResult}, it is observed that there is only 1\% increase in the overall accuracy of females as compared to males. Overall accuracy is defined as the proportion of correct kin and correct non-kin responses as compared to the total responses. 

However, from Table \ref{tab:table_humanResult}, higher $d'$ values for females as compared to males indicates higher sensitivity of females in detecting kin signal across images. This observation is also supported by the information entropy based on responses from females and males. $z$-test of proportion \cite{fleiss2004statistical} conducted at 95\% confidence level also validates this claim. These quantitative measures give us an intuition that females may have the ability to verify kin better than males. One reason for this could be that the measure being employed for testing kinship is facial similarity analogous to facial recognition; however, this needs to be tested in future studies.

%One reason for this could be that females are using facial similarity to verify kin analogous to face recognition, however this needs to be tested in future studies.

The accuracy for kinship verification increases drastically when the faces are known to the subjects. For familiar faces, female participants achieve an accuracy of 64.54\% while the male participants achieve an accuracy of 61.95\%. Also, the accuracy of non-kin verification of familiar faces is 72.47\% for females whereas it is only 52.34\% for males. This is in accordance with the belief that women perform better in episodic memory tasks \cite{Herlitz1997}. For unfamiliar faces, the trend follows the overall accuracy with females outperforming males in kinship verification. 

\subsubsection{Effect of Participant's Age on Kinship Verification}

The effect of the age of participants is studied to determine whether people of a particular age group are significantly better than others in verifying kin and non-kin. Due to limited number of participants in the younger and older age groups, the age categories have been combined into three different groups: \textless 30 years, 30-50 years, and \textgreater 50 years. As shown in Table \ref{tab:table_humanResult}, an overall accuracy of 56.95\% is observed by the participants of age-group \textgreater 50 years while the second highest accuracy is observed to be 55.6\% in the age-group of \textless 30 years. For the age group \textgreater 50, a higher $d'$ value of 0.3986 and a higher total entropy of 0.0159, as shown in Table \ref{tab:table_humanResult}, indicates that older age group may better distinguish between kin and non-kin. However, $z$-test of proportion at 95\% confidence level does not indicate statistical difference among these groups which suggests that participant's age may not have an effect on kinship verification.

\subsubsection{Effect of Stimuli Kin Pair Relation on Kinship Verification}

In a number of experiments, females have outperformed males in identifying female stimuli faces \cite{Lewin2002,Wright2003}. Therefore, it is interesting to examine if the relationship of kin pair affects the decision-making process of the participants. As shown in Table \ref{tab:table_humanResult}, the \textbf{sister-sister} kin pair has the highest overall accuracy of 66.23\%. However, using the $d'$ test of significance, it is observed that the \textbf{mother-son} pair has the highest $d'$ value of 0.8211 and the highest total entropy bits of 0.0545 as shown in Table \ref{tab:table_humanResult}. 

We also analyze the verification results separately for familiar and unfamiliar faces for different kin relations. For familiar faces, we observe that the accuracy of father-son pair increases from 53.49\% to 65.98\% and the sister-sister kin pair goes up to 82.2\% when people are familiar with the faces. This trend is seen in all the pairs and is reflective of the memory-cognitive ability of humans. As expected, the trend for unfamiliar faces is lower than familiar faces and exactly similar to the overall trend i.e. the sister-sister kin pair is the easiest to detect as kin with an accuracy of 46.0\%. 

Using the $d'$ values, it is observed that pairs having female stimuli are more accurately detected as kin. The order of the pairs based on descending $d'$ value is Mother-Son \textgreater \ Sister-Sister \textgreater \ Father-Daughter \textgreater \ Mother-Daughter \textgreater \ Brother-Sister \textgreater \ Father-Son \textgreater \ Brother-Brother. The results are in accordance with the study conducted by Kaminsky \textit{et al.} \cite{KaminskyK2009} wherein they mentioned that the presence of a female stimulus boosts the kinship accuracy. This can be attributed to partial occlusion of facial features such as beard and mustache in men as compared to women. Another reason could be the higher facial recognition capability of female participants in focusing more on female faces than male faces \cite{herlitz2013}.
%\vspace{6pt}

The results obtained for effect of participants' gender and age, as well as kin relationship between the stimuli, are used to validate our multi-approach quantitative analysis with conclusions arrived by other researchers who may not have used the same measures as we have. With this validation, we analyze the results obtained for the effect of discriminative local and global face features on kinship verification. Our motivation is to identify the top three regions from the human study to be integrated into the automatic kinship verification.  

\subsubsection{Effect of Facial Regions on Kinship Verification}

Many studies in psychology have analyzed the effect of global facial features vs. local features for face recognition abilities of humans \cite{Zhao}. Keil \cite{Keil2009} has emphasized the role of internal features in face recognition by concluding that eyes determine the optimal resolution for face recognition. These local features have been used as parts of descriptor in computational methods to verify kinship \cite{Guo2012}. However, to the best of our knowledge, no study has been conducted to analyze the effect of individual facial regions in kinship verification in a human study with statistical analysis to determine their individual effects. The two above-mentioned studies have focused on larger facial regions by dividing the face into two halves (laterally and horizontally). Intuitively, the subjects should perform better when the whole face is shown. However the results in Table \ref{tab:table_humanResult} show that even though the whole face yields an accuracy of 58.36\%, it is not very much different compared to local regions. The local features such as \textbf{not-T region} and \textbf{T region} show an accuracy of 57.02\% and 55.92\% respectively. The trend remains the same even when unfamiliar image responses are taken into account. The accuracy of T region increases to 63.45\% when the image subjects are known to humans indicating that the eye features along with the nose play an important role in kinship verification. 

These results are supported by the $d'$ test of perception and total information entropy values from the stimulus and response of participants. The complete face region has the highest $d'$ value of 0.4531 and total entropy value of 0.0221  as shown in Table \ref{tab:table_humanResult}, followed by the not-T region and the T region. A $z$-test of proportion at 95\% also validates the above pattern. The results are consistent with the face recognition studies where it has been observed that face outline, eyes, and upper face are important areas for perceiving faces \cite{Zhao}. 

%----------------------------------------------------------------------------------------------------------------------------------------

\section{Proposed Kinship Verification Learning}

\begin{figure*}
	\centering
	\includegraphics[width=.8\linewidth]{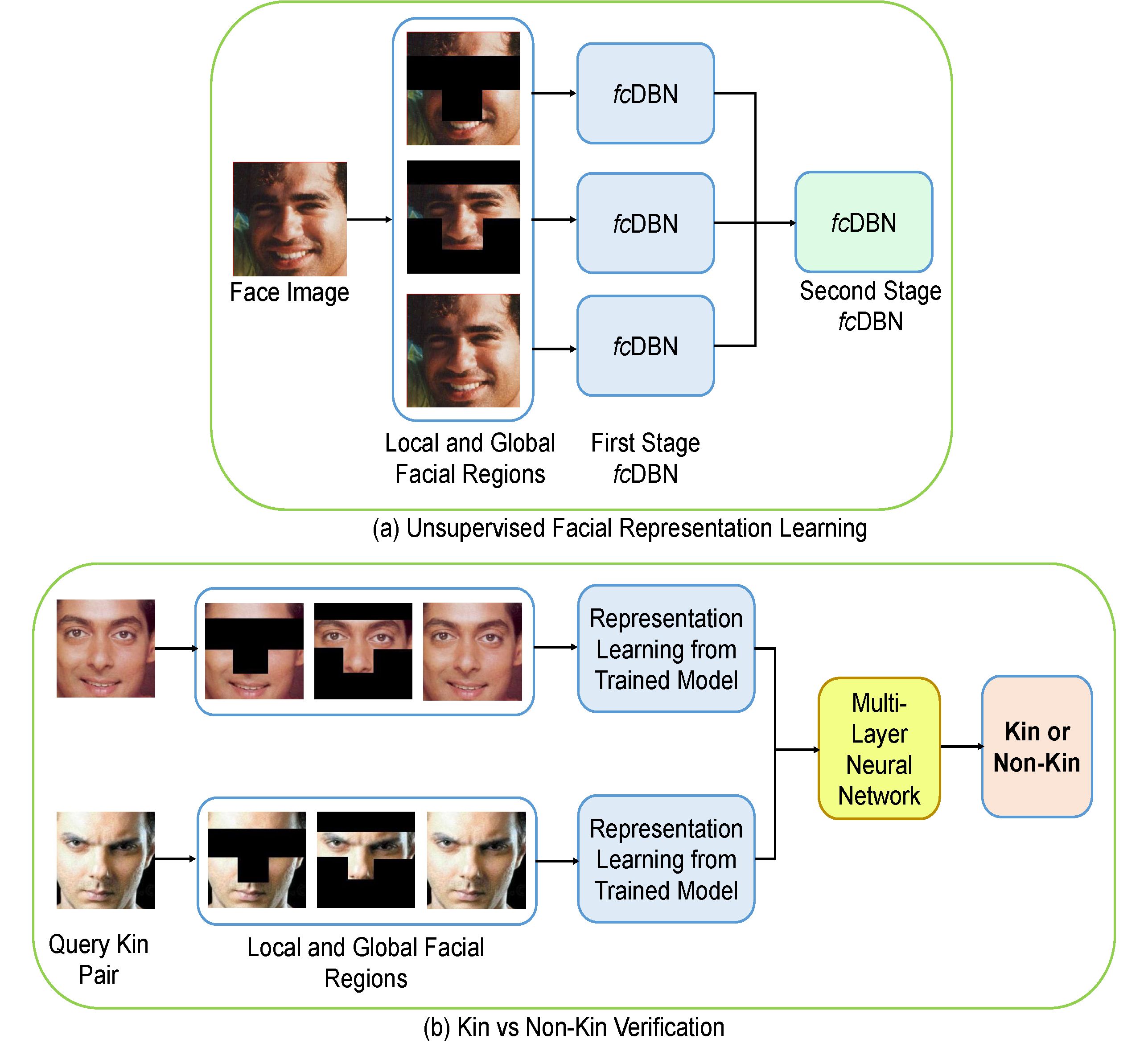}
	%\captionsetup{justification=centering}
	\caption{Proposed hierarchical kinship verification via representation learning (KVRL-\textit{fc}DBN) framework. In the first stage of Fig. 3(a), representations of individual regions are learned. A combined representation is learned in the second stage of Fig. 3(a). Fig. 3(b) shows the steps involved in kin vs non-kin classification.}
	\label{fig:Algorithm}
\end{figure*}

The analysis of human performance suggests that out of the five facial regions, full face, T-region, and not T-region yield the best performance for kinship verification. Inspired by this observation, we design a kinship verification framework that classifies a pair of input images as kin or not-kin using these three regions. As discussed earlier, it is challenging to define the similarities and differences in kin and non-kin image pairs. Therefore in this research, we propose the \textbf{Kinship Verification via Representation Learning} framework to learn the representations of faces for kinship verification using deep learning paradigm. Fig. \ref{fig:Algorithm} shows the steps involved in the proposed framework. 

In the first stage of this framework, the representations of each facial region are learned from external training data in an unsupervised manner. These are learned through the proposed \textbf{filtered contractive DBN} (\textbf{\textit{fc}DBN}) approach. The individually learned representations are combined to form a compact representation of the face in the second stage. Finally, a multi-layer neural network is trained using these learned feature representations for supervised classification of kin and non-kin. Section \ref{sec:2a} gives an overview of deep belief networks followed by the proposed filtered contractive RBMs, and Section \ref{sec:2b} describes the kinship feature learning and classification framework.

\subsection{Proposed Filtered Contractive DBN}
\label{sec:2a}

A Deep Belief Network (DBN) is a graphical model that consists of stacked Restricted Boltzmann Machines (RBM) and is trained greedily layer by layer \cite{hinton2006fast}.  An RBM represents a bipartite graph where one set of nodes is the visible layer and the other set of nodes is the hidden layer. The energy function of an RBM is defined as:

\begin{equation}
E(v,h;\theta) = - \sum_{i=1}^{D}\sum_{j=1}^{F}v_{i}W_{ij}h_{j} - \sum_{i=1}^{D}b_{i}v_{i} - \sum_{j=1}^{F}a_{j}h_{j}
\label{eq:rbm}
\end{equation}

\noindent or

\begin{equation}
E(v,h;\theta) = - \mathbf{v}^T\mathbf{W}\mathbf{h} - \mathbf{b}^T\mathbf{v} - \mathbf{a}^T\mathbf{h}
\label{eq:rbm1}
\end{equation}

\noindent{where}, $\mathbf{v} \in \{0,1\}^{D}$ denotes the visible variables and $\mathbf{h} \in \{0,1\}^{F}$ denotes the hidden variables. The model parameters are denoted by $\theta = \mathbf{\{a,b,W\}}$ and $W_{ij}$ denotes the weight of the connection between the $i^{th}$ visible unit and $j^{th}$ hidden unit and $b_{i}$ and $a_{j}$ denote the bias terms of the model. For handling real-valued visible variables such as image pixel intensities, Gaussian-Bernoulli RBMs are one of the popular formulations and the energy function is defined as:\\

\vspace{-1em}

\begin{equation}
E_r(v,h;\theta) = - \sum_{i=1}^{D}\sum_{j=1}^{F}\frac{v_{i}}{\sigma_{i}}W_{ij}h_{j} - \sum_{i=1}^{D}\frac{(v_{i}-b_{i})^2}{2\sigma^{2}} - \sum_{j=1}^{F}a_{j}h_{j}
\label{eq:gb_rbm}
\end{equation}

\noindent{Here}, $\mathbf{v} \in \mathbb{R}^{D}$ denotes the real-valued visible vector and $\theta=\mathbf{\{a,b,W,\sigma\}}$ are the model parameters. The joint distribution over $\mathbf{v}$ and $\mathbf{h}$, and the marginal distribution over $\mathbf{v}$ is defined as:

\begin{equation}
P(v,h) = \frac{1}{Z} exp(-E(v,h;\theta))
\end{equation}

and

\begin{equation}
P(v) = \sum_{h}P(v,h)
\end{equation}

where, $Z = \sum_{v,h} exp(-E(v,h))$ is a partition function.

Let $L_{RBM}$ be the loss function of RBM with the energy function defined in Eq. \ref{eq:rbm1}. It can be defined as

\begin{equation}
 L_{RBM} = - \sum_{i=1}^{n} \log P(v_{i})
 \label{loss_rbm}
\end{equation}

In this paper, we extend this formulation and propose filtered contractive DBN (\textit{fc}DBN) which utilizes filtered contractive RBMs (\textit{fc}RBM) as its building block. \textit{fc}RBM has two components: a contractive regularization term and a filtering component which is discussed in detail below.

The idea of introducing contractive penalty stems from Rifai et. al \cite{rifai2011contractive} where they introduce contractive autoencoders. A regularization term is added in the autoencoder loss function for learning robust features as shown in Eq. \ref{eq:contractiveautoencoder}. 
 
\begin{equation}
\begin{split}
L_{AE} & = \mathbf{arg \ min_{\theta}} \parallel \mathbf{v} - \phi (\bm{\mathcal{W}}' ( \phi( \bm{\mathcal{W}} \mathbf{v} + b) ) + b') \parallel^{2} \\
& + \lambda \parallel \mathcal{J}\parallel^2_F \\
\end{split}
\label{eq:contractiveautoencoder}
\end{equation}

\noindent where, $\theta=\{\bm{\mathcal{W}},b\}$ represents the weight and the bias of the autoencoder to be learned, $\phi$ represents the activation function, $\lambda$ represents the regularization parameter, and  \[  \parallel\mathcal{J}\parallel^2_F \  = \ \parallel(J(\phi(\bm{\mathcal{W}}\mathbf{v}))) \parallel^{2}_{F} \] represents the Jacobian of the input with respect to the encoder function of the autoencoder. For a linear activation function, the contractive penalty boils down to a simple weight decay term (Tikhonov-type regularization). For a sigmoid the penalty is smooth and is given by:

\begin{equation} 
\begin{split}
\parallel\mathcal{J}\parallel^2_F &= \parallel J ( \phi ( \bm{\mathcal{W}} \mathbf{v} ) ) \parallel^{2}_{F} \\ & = \sum_i \Big( \phi(\bm{\mathcal{W}} \mathbf{v})_{i} (1 - \phi( \bm{\mathcal{W}} \mathbf{v})_{i} )  \Big)^2 \sum_j \bm{\mathcal{W}}_{ij}^2  
\end{split}
\label{eq:jacobian}
\end{equation}

Our work is motivated by the analytic insight and practical success of contractive autoencoders. We propose to apply the contractive penalty term to the RBM formulation. Thus, the modified loss function for contractive RBMs (\textbf{c-RBM}) can be expressed as:

\begin{equation}
\mathcal{L}_{c-RBM} =  L_{RBM} + \alpha \parallel {\mathcal{J}} \parallel^{2}_F 
\label{eq:norm2}
\end{equation}

\noindent where, $\parallel \mathcal{J} \parallel^{2}_F $ represents Frobenius norm of the Jacobian matrix (i.e. it is $l_2$-norm of the second order differential) as shown in Eq. \ref{eq:jacobian}. Penalizing the Frobenius norm of the Jacobian matrix leads to penalization of the sensitivity; which encourages robustness of the representation. The contractive penalty encourages the mapping to the feature space to be contractive to the neighborhood of the training data. The flatness induced by having low valued first derivatives will lead to invariance of the representation for small variations in the input.

We further introduce a filtering approach in the RBM. Facial images have an inherent structure and filters can be used to extract this structural information in order to train the network using only the relevant filtered information. Therefore, we propose extending Eq. \ref{eq:rbm1} (and in a similar manner, Eq. \ref{eq:gb_rbm}) with a filtering approach that can incorporate the structural and relational information in the image using filters. 

\begin{equation}
E_{f}(\mathcal{V}_{k},h;\theta_f) = - \mathcal{V}_{k} ^T\mathbf{W}\mathbf{h} - \mathbf{b}^T \mathcal{V}_{k} - \mathbf{a}^T\mathbf{h}
\label{eq:rbm_new}
\end{equation}

\noindent where, $\mathcal{V}_{k} =  \mathbf{(f}_{k} \cdot \mathbf{v})$ and ``$\cdot$'' is the convolution operation. $\mathbf{f}_k$ is the $k^{th}$ learned filter of size $mn$ and therefore, $\theta_{f}$ includes $\mathbf{f}_k$ and other weight parameters. Here, the filters $\mathbf{f}_k$  transform the input image $\mathbf{v}$, emphasizing relevant structural information which is used to train the RBM. Utilizing the above energy function, the loss function of the filtered RBM, $L_{fRBM}$ is defined similarly to Eq. \ref{loss_rbm}. Note that, the proposed formulation is different from convolutional RBMs \cite{lee2009convolutional}. In convolutional RBMs, the weights are shared among all locations in the image and thus, a pooling step is required to learn high-level representations. In the proposed formulation, we have introduced separate filters that will account for the structure of the image and learn these filters and weight matrix simultaneously.

Combining the above two components, we define filtered contractive RBMs (\textit{fc}RBM) and the loss function is modeled as:

\begin{equation}
\mathcal{L}_{\textit{fc}RBM} =  L_{fRBM} + \alpha \parallel {\mathcal{J}} \parallel^{2}_F +  \beta \parallel \mathbf{f} \parallel^2_{2} 
\label{eq:norm3}
\end{equation}

\noindent where, $\alpha$ and $\beta$ are the regularization parameters. $l_{2}$-norm applied over the filters prevents large deviation of values that could potentially have an unwarranted filtering effect on the images. Both the components of the proposed formulation are smooth and hence differentiable; and can be solved iteratively using contrastive divergence based approach. Multiple \textit{fc}RBMs are then stacked together to form \textit{fc}DBN.

\subsection{KVRL-\textit{fc}DBN for Kinship Verification}
\label{sec:2b}

The KVRL framework proposed in this research comprises of two phases: 
\begin{itemize} 
	\item Unsupervised hierarchical two-stage face feature representation learning 
	\item Supervised training using extracted features and kin verification using the learned model
\end{itemize} 

%There are two versions of the proposed KVRL framework - one with stacked denoising autoencoders (SDAE) and one with deep belief networks (DBN). %Among many deep learning algorithms, SDAE and DBN are two very popular techniques.

\textbf{KVRL-\textit{fc}DBN:} The representation of face image is learned by stacking \textit{fc}RBMs and learning the weights in a greedy layer by layer fashion to form a filtered contractive deep belief network (\textit{fc}DBN). As shown in Fig. \ref{fig:Algorithm}, we extract three regions from the input face image to learn both global and local features. These regions are selected based on the results of the human study that indicates complete face, T region and not-T region are more significant than other face regions. In the first stage of the proposed KVRL-\textit{fc}DBN framework, each region is first resized to a standard $M \times N$ image and is converted to $1 \times MN$ vector. Three separate \textit{fc}DBNs are trained, one for each region and the output from these \textit{fc}DBNs are combined using another \textit{fc}DBNs which acts as the second stage of the proposed hierarchical feature learning.

We next apply \textit{dropout} based regularization throughout the architecture. Srivastava et al. \cite{hinton2014} proposed \textit{dropout} training as a successful way for preventing overfitting and an alternate method for regularization in the network. The motivation is to inhibit the complex co-adaptation between the hidden nodes by randomly dropping out a few neurons during the training phase. It can be seen as a sampling process from a larger network to create random sub-networks with the aim of achieving good generalization capability. Let $f$ denote the activation function for the $n^{th}$ layer, and $\mathbf{W}$, $\mathbf{b}$ be the weights and biases for the layer, $*$ denotes the element-wise multiplication, and $\mathbf{m}$ is a binary mask with entries drawn $i.i.d.$ from \textit{Bernoulli} (1-r) indicating which activations are not dropped out. Then the forward propagation to compute the activation $\mathbf{y_{n}}$ of $n^{th}$ layer of the architecture involving dropout can be calculated as,

\begin{equation}
y_{n} = f \left( \frac{1}{1-r} y_{n-1} * \mathbf{mW} + \mathbf{b} \right)
\end{equation}

By introducing \textit{dropout} in the proposed approach, we obtain good generalization that emulates sparse representations to mitigate any possible overfitting. In summary, while the first stage of the KVRL-\textit{fc}DBN framework learns the local and global facial features, the second stage assimilates the information (i.e. feature fusion) which is used for kinship verification.

\begin{figure*}[!t]
	\centering
	\includegraphics[width=.95\linewidth]{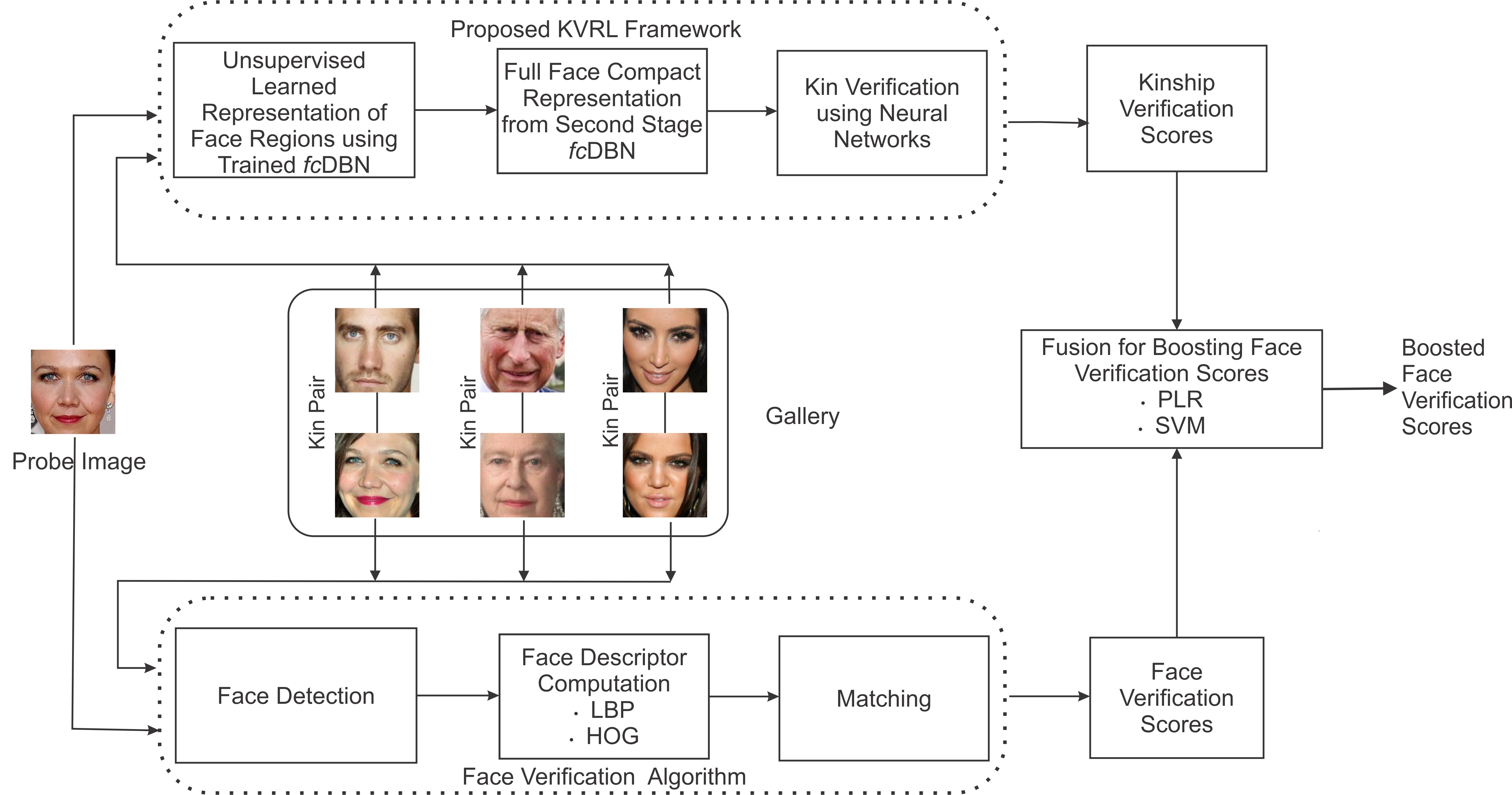}
	\caption{Illustrating the steps involved in the proposed context boosting algorithm where kinship verification scores generated from the KVRL framework are used to improve the face verification performance.}
	\label{fig:KinAided}
\end{figure*}

\begin{figure}[!t]
	\centering
	\includegraphics[width=.99\linewidth]{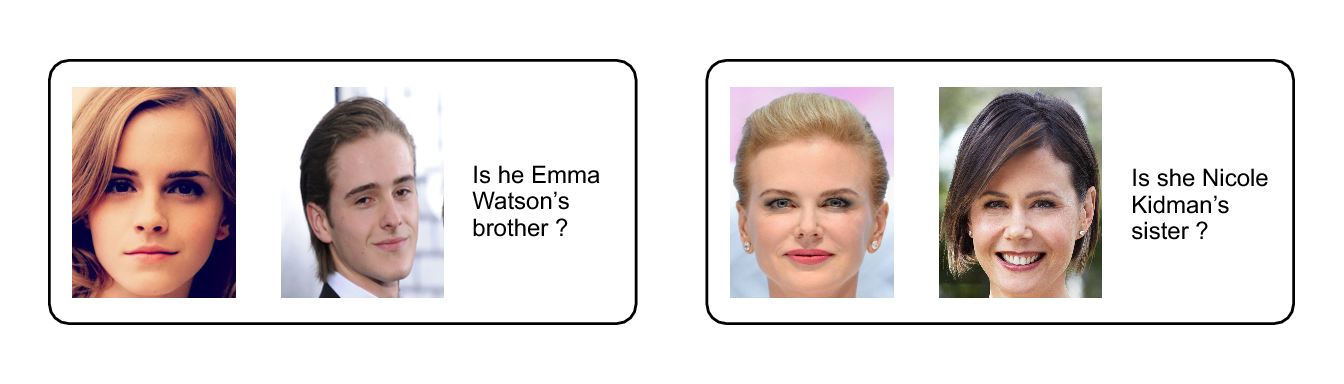}
	%\captionsetup{justification=centering}
	\vspace{-25pt}
	\caption{Humans utilize kinship as a context to identify siblings of famous personalities.}
	\label{fig:ContextBoosting}
\end{figure} 

The number of images in currently available kinship datasets are limited and cannot be used directly to train the deep learning algorithms. Therefore, a separate database is needed to train the model employed in the KVRL-\textit{fc}DBN framework (details are given in Section \ref{secv:a}). The representations learned from the proposed KVRL-\textit{fc}DBN framework are used for kinship verification. As shown in Fig. \ref{fig:Algorithm}(b), for a pair of kin images, the features are concatenated to form the input vector for supervised classification. A three-layer feed-forward neural network is trained for classifying the image pair as kin or non-kin. 

\section{Boosting Face Verification using Kinship}

Soft biometrics modalities lack the individualization characteristics on their own but can be integrated within a verification system that uses the primary biometric trait such as face to boost the accuracy \cite{jain2004soft}. Soft biometric traits can often be based on association wherein the context of association can be used to increase the recognition performance in challenging image scenarios \cite{bharadwaj2014}. In this research, we propose kinship as a \textit{context} that can be used as a soft biometric modality to improve the accuracy of face verification. Kinship cues are used by humans in daily life for recognition. For instance, we may recognize a person based on their familiarity with their kin even though we may not have met the person earlier. Such a scenario is depicted in Fig. \ref{fig:ContextBoosting}. To incorporate this context, we propose a formulation to incorporate kinship verification scores generated by the proposed framework to boost the performance of any face verification algorithm.

\begin{figure}[!t]
	\centering
	\includegraphics[width=.6\linewidth]{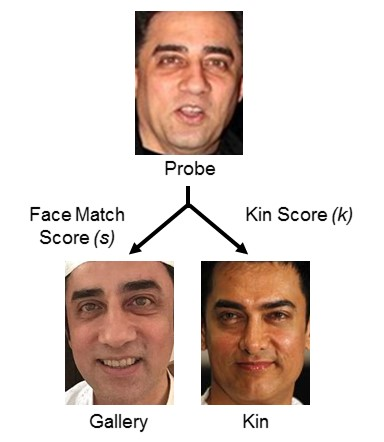}
	%\captionsetup{justification=centering}
	\caption{A probe image can have a match score (\textit{s}) with an image in the gallery and a kin score (\textit{k}) with the associated kin in the gallery to boost the face verification performance.}
	\label{fig:KinSoft}
\end{figure}

Fig. \ref{fig:KinAided} shows how the proposed KVRL-\textit{fc}DBN framework is used to improve the performance of face verification algorithms using kin-verification scores. This formulation is generic in nature and independent of the kinship verification and face verification algorithms. As shown in Fig. \ref{fig:KinSoft}, given a probe face image, face verification score and kinship classification score are computed from the gallery data (claimed identity and associated kin image), which are then used in the proposed formulation. We demonstrate two methods for boosting the performance using Product of Likelihood Ratio (PLR) \cite{plr} and Support Vector Machine (SVM) \cite{vapnik1998statistical}.

\begin{itemize}

\item \textbf{PLR based Score Boosting Algorithm}: Let $s$ be the face matching score obtained by matching a probe image and a gallery image. $k_{1},k_{2},\ldots,k_{n}$ represent the kin scores obtained from the probe image and images of the gallery subject. The product of likelihood ratio \cite{nanda2008} can be calculated as:

\begin{equation}
	PLR = \frac{P(s\mid \omega_{1})}{ P( s \mid \omega_{2} )} \times \prod_{i=1}^{N} \frac{P(k_{i} \mid ks_{1})}{ P( k_{i} \mid ks_{2} )} 
\end{equation}

\noindent Here, $ks_{1}$ represents the true kin class, $ks_{2}$ represents the non-kin class, $\omega_{1}$ represents the genuine class, $\omega_{2}$ represents the impostor class. $P(s\mid\omega_{1})$ and $P(s\mid\omega_{2})$  represent the class conditional probability of the input vector. All four variables are modeled using mixture of Gaussian distributions.

\item \textbf{SVM based Score Boosting Algorithm:} Let $\mathbf{p_{i}}$ be the feature vector representing the concatenation of the face matching and kin verification scores i.e $\mathbf{p_{i}} = [ s_{i}  \ k_{i} ] $. A support vector machine can be trained on the combined score vector to boost the performance of face verification.

\end{itemize}

Since we are proposing a generic approach which is independent of the features used for face verification, we have used the commonly explored local binary patterns (LBP) \cite{ahonen2006face} and histogram of oriented gradients (HOG) \cite{dalal2005histograms} for face verification. 

%-----------------------------------------------------------------------------------------

\section{Experimental Evaluation}

This section describes the datasets, implementation details, and experimental protocols used for evaluating the effectiveness of the proposed representation learning for kinship using hierarchical multi-stage filtered contractive deep belief network (KVRL-\textit{fc}DBN) along with the PRL and SVM based face verification score boosting algorithms. 

\begin{table*}[t]
\centering
\captionsetup{justification=centering}
\caption{Characteristics of the five databases used in this research.}
\label{tab:dba}
\begin{tabular}{|l|c|c|c|c|}
\hline
{\bf Database}    & {\bf No. of Subjects} & {\bf Total Images} & {\bf Kin Relations} & {\bf Multiple Images} \\ \hline
Cornell Kin \cite{Fang2010}      & 286                   & 286                       & 4                          & No                                                                           \\ \hline
UB KinFace \cite{Buffalo_TMM_Kinship}      & 400                   & 600                       & 4                          & No                                                                           \\ \hline
KinFaceW-I \cite{lu}       & 1066                  & 1066                      & 4                          & No                                                                           \\ \hline
KinFaceW-II  \cite{lu}     & 2000                  & 2000                      & 4                          & No                                                                           \\ \hline
WVU Kinship & 226             & 904                 & 7                    & Yes                                                                    \\ \hline
\end{tabular}
\end{table*}

\subsection{Datasets}
\label{secv:a}

The efficacy of the proposed kinship verification algorithm is evaluated on the following four publicly available databases. 
\begin{itemize}
	\item UB KinFace Dataset\cite{Buffalo_TMM_Kinship},
	\item Cornell Kinship Dataset \cite{Fang2010},
	\item KinFace-I \cite{lu}, and
	\item KinFace-II \cite{lu}.   
\end{itemize} 

\begin{figure}[!t]
	\centering
		\includegraphics[width=.9\linewidth]{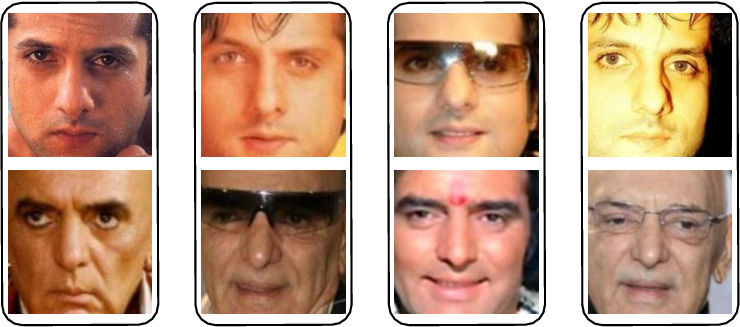}
	%\captionsetup{justification=centering}
	\caption{Challenges of pose, illumination, and occlusion in multiple images of the same kin-pair.}
	\label{fig:KinChallenges}
\end{figure}

Along with these four, we have also prepared a new kinship database, known as the \textit{WVU Kinship Database}, containing multiple images of every person\footnote{The chrominance based algorithm, given by Bordallo et al. \cite{comments2015} performs poorly on the WVU Kinship database which validates the correctness of the database.}. The WVU Kinship dataset consists of 113 pairs of individuals. The dataset has four images per person, which allows us to have intra-class variations for a specific kin-pair along with the inter-class variations  generally available with all other databases. It consists of seven kin-relations: Brother-Brother (BB), Brother-Sister (BS), Sister-Sister (SS), Mother-Daughter (MD), Mother-Son (MS), Father-Son (FS), and Father-Daughter (FD). The database has 22 pairs of BB, 9 pairs of BS, 13 pairs of SS, 14 pairs of FD, 34 pairs of FS, 13 pairs of MD and 8 pairs of MS where every pair has eight images each. As shown in Fig. \ref{fig:KinChallenges}, the multiple images per kin-pair also include variations in pose, illumination and occlusion. Table \ref{tab:dba} summarizes the characteristics of all five databases.

Kinship verification results are shown on all five databases. However, the results of face score boosting are shown only on the WVU Kinship database because the other four databases only contain a single image per person.

\begin{figure*}[!t]
	\centering
	\begin{subfigure}{.33\linewidth}
		\includegraphics[width=.95\linewidth]{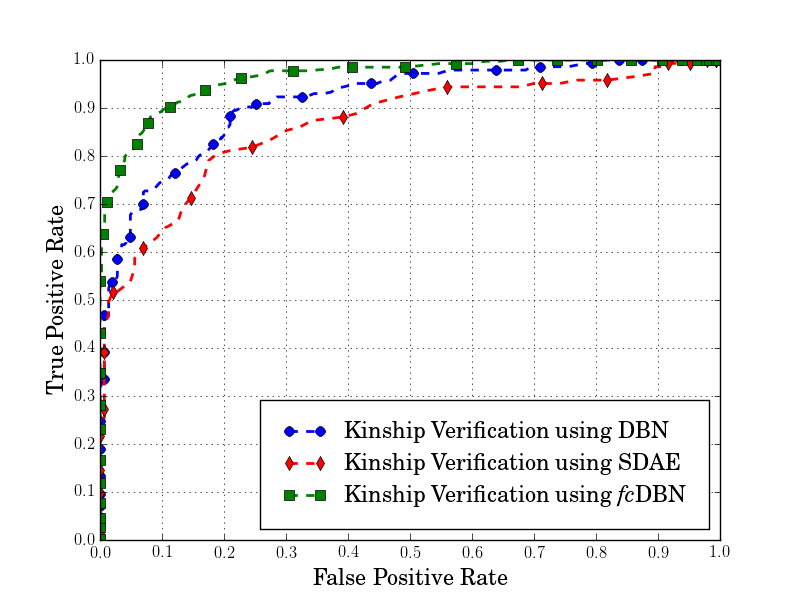}
		\label{Fig_Cornell}
		%\captionsetup{justification=centering}
		\caption{Cornell Kinship Database}
		\vspace*{5mm}
	\end{subfigure}%
	\begin{subfigure}{.33\linewidth}
		\includegraphics[width=.95\linewidth]{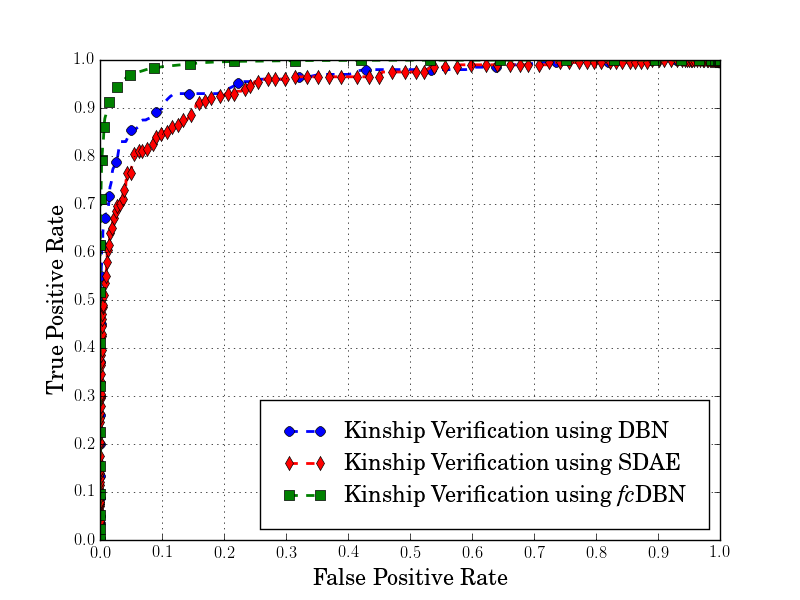}
		\label{Fig_KinfaceI}
		%\captionsetup{justification=centering}
		\caption{KinFace-I Database}
		\vspace*{5mm}
	\end{subfigure}%
	\begin{subfigure}{.33\linewidth}
		\includegraphics[width=.95\linewidth]{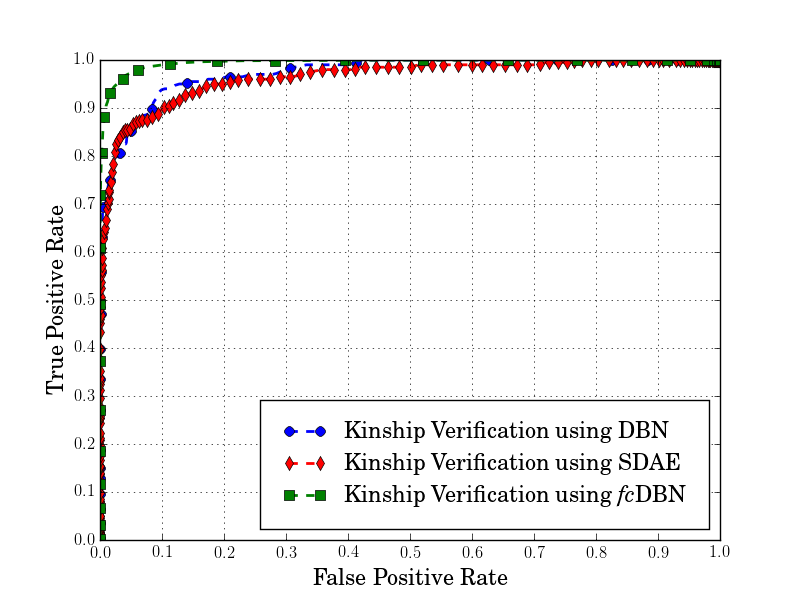}
		\label{Fig_KinfaceII}
		%\captionsetup{justification=centering}
		\caption{KinFace-II Database}
		\vspace*{5mm}
	\end{subfigure} 
	\begin{subfigure}{.33\linewidth}
		\includegraphics[width=.95\linewidth]{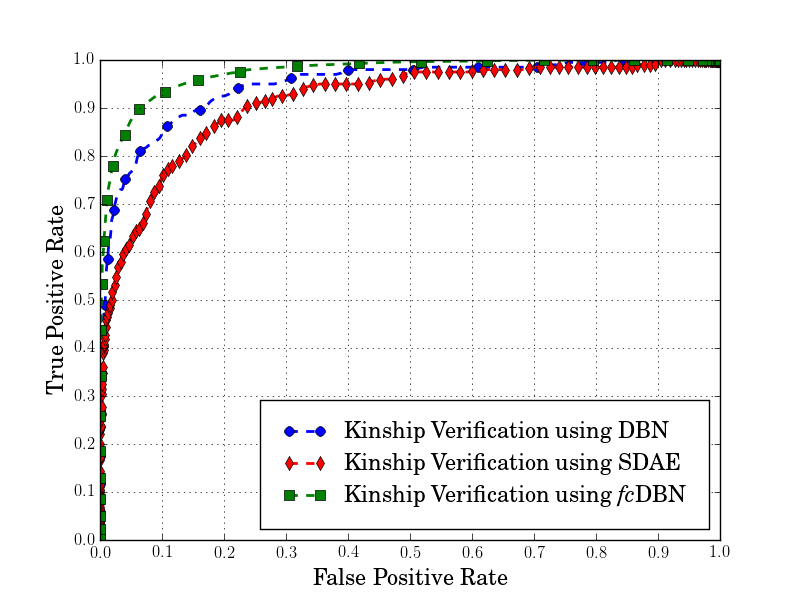}
		\label{Fig_UBI}
		%\captionsetup{justification=centering}
		\caption{UB Kinship Database}
	\end{subfigure}
	%	\begin{subfigure}{.33\linewidth}
	%	\includegraphics[width=.95\linewidth]{Fig_Kin_UB2}
	%	\label{Fig_UBII}
	%	\captionsetup{justification=centering}
	%	\caption{UB-Old Parent and Young Child Database}
	%	\end{subfigure}%
	\begin{subfigure}{.33\linewidth}
		\includegraphics[width=.95\linewidth]{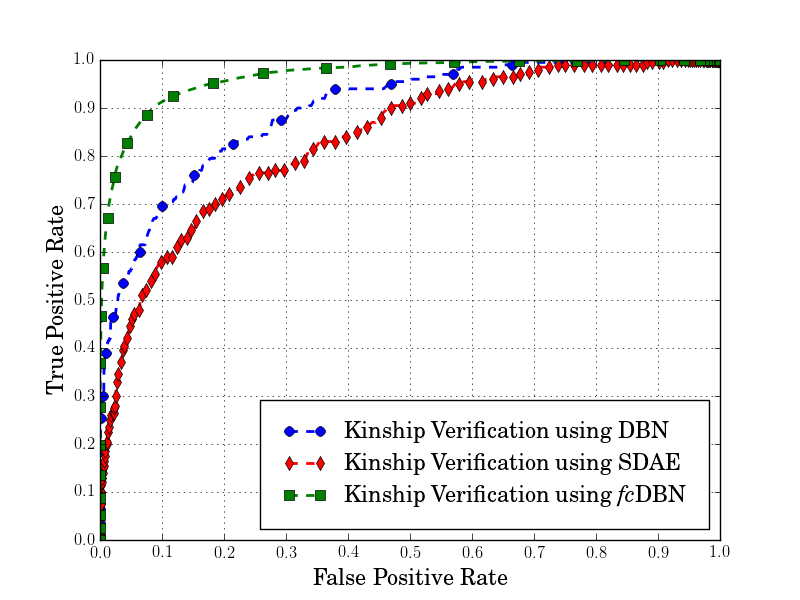}
		\label{Fig_WVU}
		%\captionsetup{justification=centering}
		\caption{WVU Kinship Database}
	\end{subfigure}
	\captionsetup{justification=centering}
	\caption{Results of kinship verification using the proposed hierarchical KVRL framework.}
	\label{fig:kin_roc}
\end{figure*}

\subsection{Implementation Details}

Training the \textit{fc}DBN algorithm to learn the face representation for kinship requires a large number of face images. For this purpose, about 600,000 face images are used. These images are obtained from various sources including CMU-MultiPIE and Youtube faces databases \cite{wolf2011}, \cite{MultiPie2010}. Note that, existing algorithms do not use outside data; however, as mentioned previously, due to the nature of deep learning paradigm, the proposed algorithm requires large data to learn face representation useful for kinship verification. %It is important to note that this is not a requirement for the results reported in the literature. 

For face detection, all the images are aligned using affine transformation and Viola-Jones face detection algorithm \cite{viola2004robust}. Facial regions are extracted from each image and resized to $32 \times 32$. The resized regions are converted to a vector of size $1024$ and given as input to individual \textit{fc}DBN deep learning algorithm in the first stage of Fig \ref{fig:Algorithm}(a). For every individual \textit{fc}DBN, three filtered contractive RBMs are stacked together and all of them are learned in a greedy layer-wise fashion where each layer receives the representation of the output from the previous layer. In the first stage, the number of nodes are 1024, 512, and 512 respectively. An output vector of size 512 is obtained from each deep belief network and is concatenated to form a vector of size 1536. A compact representation is learned from the \textit{fc}DBN in the second stage and is used for training the classifier. In the second stage of the deep belief network, the size of the three layers are 1536, 1024, and 512, respectively. The \textit{dropout} is applied with probability $0.5$ on the hidden nodes and $0.2$ on the input vectors. The performance of the proposed KVRL-\textit{fc}DBN algorithm is also evaluated when only face is used or all the five facial regions (shown in Fig. \ref{fig:ExampleOfFaceRegions}) are used. 
%Since the architecture is flexible in nature, we also compare the results using both Sparse Denoising Autoencoders (SDAE) and Deep Belief Network (DBN) in the KVRL framework. We term these approaches of KVRL framework as KVRL-SDAE and KVRL-DBN. Similar process is followed for KVRL-DBN and KVRL-SDAE. For every individual SDAE, three different autoencoders are stacked together and all of them are learned For all three facial regions, the number of nodes in the SDAE learned in the first stage is same $\left[AE^{1}_{j}, AE^{2}_{j}, AE^{3}_{j}\right] = [1024, 512, 256]$.  An output vector of size $256$ is obtained from the third level of each stacked denoising autoencoder. These outputs are the representations learned from the global and local face regions, and are concatenated to form a vector of size $768$. This vector is input to the stacked denoising autoencoder in the second stage of the proposed KVRL architecture which yields the final vector of size $512$. This is the final learned representation of the \emph{face} image and used for classification with neural network. 

\begin{table*}[!h]
	\small
	\centering
	%\begingroup
	\begin{tabular}{|L{7em}|c|c|c|c|c|}
		\hline
		\textbf{Algorithm}  & \textbf{Cornell} & \textbf{UB} & \textbf{KinFace-I} & \textbf{KinFace-II} &\textbf{WVU} \\ \hline
		
		KVRL-SDAE &  82.0                                  & 85.9                            &92.3                                    & 92.7                                     & 78.7                             \\ \hline
		KVRL-DBN                         & 83.6                                 & 88.3                           & 93.0                                   & 93.9                                     & 83.5                              \\ \hline
		\textbf{KVRL-\textit{fc}DBN }                                        & \textbf{89.5}                                 & \textbf{91.8}                           & \textbf{96.1}                                   &\textbf{96.2}                                    & \textbf{90.8}                              \\ \hline
	\end{tabular}
	%\endgroup
	\captionsetup{justification=centering}
	\caption{Kinship verification performance of the proposed KVRL framework on 5 different datasets}
	\label{tab:kin_algo_table}
	%\vspace{6pt}
\end{table*}
\begin{table*}[!htp]
	\hspace{-0.5cm}
	\small
	\begin{subtable}{.5\linewidth}
		\centering
		\caption{Cornell Kinship Dataset}
		\begin{tabular}{|l|c|c|c|c|}
			\hline 
			Algorithm & FS & FD & MS & MD \\ \hline
			MNRML\cite{lu} & 74.5 & 68.8 & 77.2 & 65.8 \\ \hline
			DMML\cite{Yan} & 76.0 & 70.5 & 77.5 & 71.0 \\ \hline
			KVRL using SDAE    &   85.0 &  80.0  &  85.0  &   75.0 \\ \hline
			KVRL using DBN    &  88.3  &  80.0  &  90.0  & 72.5  \\ \hline
			KVRL using c-DBN    &  90.0  &  84.8  &  90.0  & 78.9  \\ \hline
			\textbf{KVRL using \textit{fc}DBN}    &  91.7  &  87.9  &  95.2  & 84.2  \\ \hline
		\end{tabular}
		\label{tab:res1}
	\end{subtable} \hspace{0.5mm} %
	\begin{subtable}{.5\linewidth}
		\centering
		\caption{UB Kinship Dataset}
		\begin{tabular}{|l|c|c|}
			\hline 
			Algorithm & Child-Young Parents & Child-Old Parents \\ \hline
			MNRML\cite{lu} & 66.5 & 65.5 \\ \hline
			DMML\cite{Yan} & 74.5 & 70.0 \\ \hline
			KVRL using SDAE    &   85.9 &  84.8 \\ \hline
			KVRL using DBN    &   88.5 &  88.0    \\ \hline
			KVRL using c-DBN    &  90.0  &  89.5 \\ \hline
			\textbf{KVRL using \textit{fc}DBN}    &  92.0  &  91.5 \\ \hline
		\end{tabular}
		
		\label{tab:res2}
	\end{subtable}\\[2ex]
	\hspace{-0.5cm}
	\begin{subtable}{.5\linewidth}
		\centering
		\caption{KinFace-I Dataset}
		\begin{tabular}{|l|c|c|c|c|}
			\hline 
			Algorithm & FS & FD & MS & MD \\ \hline
			MRNML\cite{lu} & 72.5 & 66.5 & 66.2 & 72.0 \\ \hline
			DML\cite{Yan} & 74.5 & 69.5 & 69.5 & 75.5 \\ \hline
			Discriminative\cite{dehghanOVS14} & 76.4 & 72.5 & 71.9 & 77.3 \\ \hline
			KVRL using SDAE    &   95.5 &  88.8  &  87.1  &   96.9 \\ \hline
			KVRL using DBN    &   96.2 &  89.6  & 87.9  &  97.6 \\ \hline
			KVRL using c-DBN    &  97.4  &  93.3  &  90.5  & 98.4  \\ \hline
			\textbf{KVRL using \textit{fc}DBN}    &  98.1  &  96.3  &  90.5  & 98.4  \\ \hline
		\end{tabular}
		\label{tab:res3}
	\end{subtable} \hspace{0.5mm} %
	\begin{subtable}{.5\linewidth}
		\centering
		\caption{KinFace-II Dataset}
		\begin{tabular}{|l|c|c|c|c|}
			\hline 
			Algorithm & FS & FD & MS & MD \\ \hline
			MNRML\cite{lu} & 76.9 & 74.3 & 77.4 & 77.6 \\ \hline
			DML\cite{Yan} & 78.5 & 76.5 & 78.5 & 79.5 \\ \hline
			Discriminative\cite{dehghanOVS14} & 83.9 & 76.7 & 83.4 & 84.8 \\ \hline
			KVRL using SDAE   &   94.0 &  89.2  &  93.6  &   94.0 \\ \hline
			KVRL using DBN    &  94.8  &  90.8  &  94.8 & 95.6  \\ \hline
			KVRL using c-DBN   &  96.0  &  92.4  &  96.4  & 96.8  \\ \hline
			\textbf{KVRL using \textit{fc}DBN}    &  96.8  &  94.0  &  97.2  & 96.8  \\ \hline
		\end{tabular}
		\label{tab:res4}
	\end{subtable}\\[2ex]
	\begin{subtable}{\linewidth}
		\centering
		\caption{WVU Kinship Dataset}
		\begin{tabular}{|l|c|c|c|c|c|c|c|}
			\hline 
			Algorithm & FS & FD & MS & MD & BB & BS & SS \\ \hline
			KVRL using SDAE    &   80.9 &  76.1  &  74.2  &   80.7 & 81.6 & 76.5 & 80.3\\ \hline
			KVRL using DBN    &  85.9  &  79.3  & 76.0  &  84.8 & 85.0 & 79.9 & 85.7 \\ \hline
			KVRL using c-DBN    &  87.9  &  79.9  &  83.6  & 91.3 & 86.9 & 82.6 & 91.8  \\ \hline
			\textbf{KVRL using \textit{fc}DBN}    &  90.8  &  84.4  &  90.6  & 95.2 & 90.9 & 87.5 & 95.7  \\ \hline
		\end{tabular}
		\label{tab:res5}
	\end{subtable}
	%\captionsetup{justification=centering}
	\caption{Comparing the kinship verification performance (\%) of the proposed KVRL framework with existing kinship verification algorithms on multiple datasets.}
	\label{tab:kin_table}
	
\end{table*}

\begin{figure*}[!ht]
	\centering
	\begin{subfigure}{.5\linewidth}
		\includegraphics[width=0.95\linewidth]{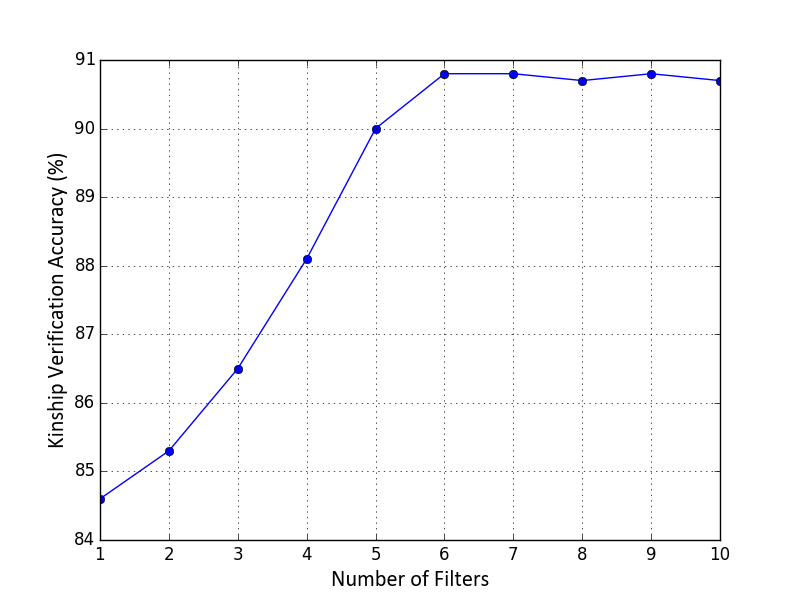}
		\label{Filters}
		%\captionsetup{justification=centering}
		\caption{Verification performance with changing the number of filters.}
	\end{subfigure}%
	\begin{subfigure}{.5\linewidth}
		\includegraphics[width=.95\linewidth]{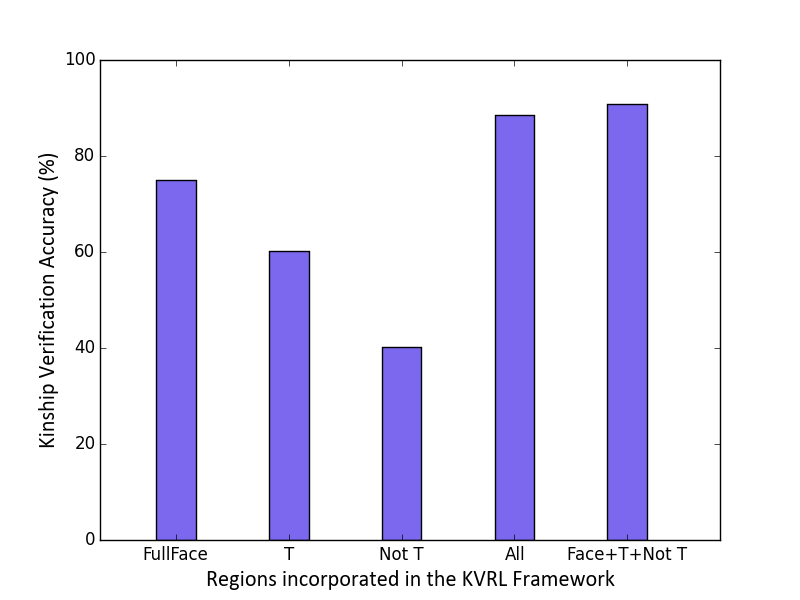}
		\label{barplot}
		%\captionsetup{justification=centering}
		\caption{Kinship verification performance with respect to regions taken in the KVRL-\textit{fc}DBN framework.}
	\end{subfigure}
	\captionsetup{justification=centering}
	\caption{Variations in the performance of KVRL-\textit{fc}DBN with respect to number of filters and type of facial regions on the WVU kinship database.}
	\label{fig:Fig_Kinship_Results}
\end{figure*}%
\subsection{Experimental Protocol} 

\subsubsection{Kinship Verification} 

The performance of the proposed KVRL-\textit{fc}DBN framework is evaluated on the same experimental protocol as described by Yan et al. \cite{Yan}, where five-fold cross-validation for kin verification is performed by keeping the images in all relations to be roughly equal in all folds. This protocol is followed to ensure that the experimental results are directly comparable even though the list of negative pairs may vary. In this algorithm, a random negative pair for kinship is generated such that each image is used only once in the training phase. 
\renewcommand*{\thefootnote}{\fnsymbol{footnote}}
The performance of the proposed algorithm (KVRL-\textit{fc}DBN) is compared with the baseline evaluations of KVRL framework along with three state-of-the-art algorithms.
\begin{itemize}

\item Multiview neighborhood repulsed metric learning (MNRML)\cite{lu}\footnote{\label{note1}Since the experimental protocol is same, results are directly reported from the papers.}, 
\item Discriminative multi-metric learning (DMML)\cite{Yan}\footref{note1} , and 
\item Discriminative model\cite{dehghanOVS14}\footref{note1} . 
\end{itemize}

Since the proposed architecture is flexible in nature, we also utilize Sparse Denoising Autoencoders (SDAE) and Deep Belief Network (DBN) in the KVRL framework. We term these approaches of KVRL framework as KVRL-SDAE and KVRL-DBN. The proposed approach (KVRL-\textit{fc}DBN) is compared with KVRL-SDAE, KVRL-DBN and KVRL-cDBN (where contractive RBMs are utilized in the KVRL framework). We also analyze the effect of regions is observed where different combinations of facial regions are given as input to the KVRL-\textit{fc}DBN framework.

\subsubsection{Boosting Face Verification using Kinship as Context}

The WVU Kinship database is divided into training and testing sets. Similar to kinship verification experiments, the training partition consists of 60\% of the dataset and the testing partition consists of the remaining 40\% where the subjects are mutually independent and disjoint. In both the sets, two images of an individual are used as probe, while the remaining are used as gallery. Four images of the kin of the individual are kept in the gallery where the association between the kin in the gallery set is known. The proposed KVRL-\textit{fc}DBN framework is used to generate the kinship scores between the probes and kin images using the \textit{fc}DBN deep learning algorithm.

\begin{figure*}[!t]
	\centering
	\begin{subfigure}{.5\linewidth}
	\includegraphics[width=0.95\linewidth]{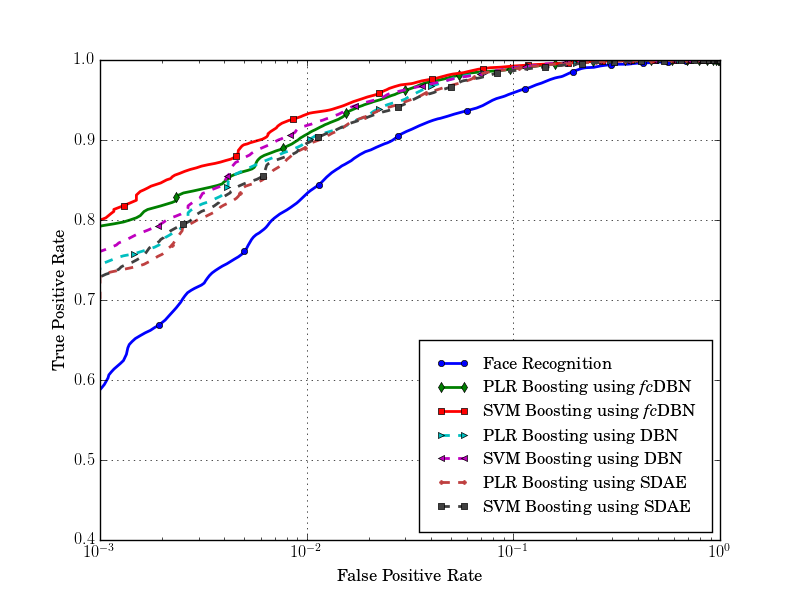}
	\label{Fig_Aided_HOG}
	%\captionsetup{justification=centering}
	\caption{ROC using HOG descriptor}
	\end{subfigure}%
	\begin{subfigure}{.5\linewidth}
	\includegraphics[width=.95\linewidth]{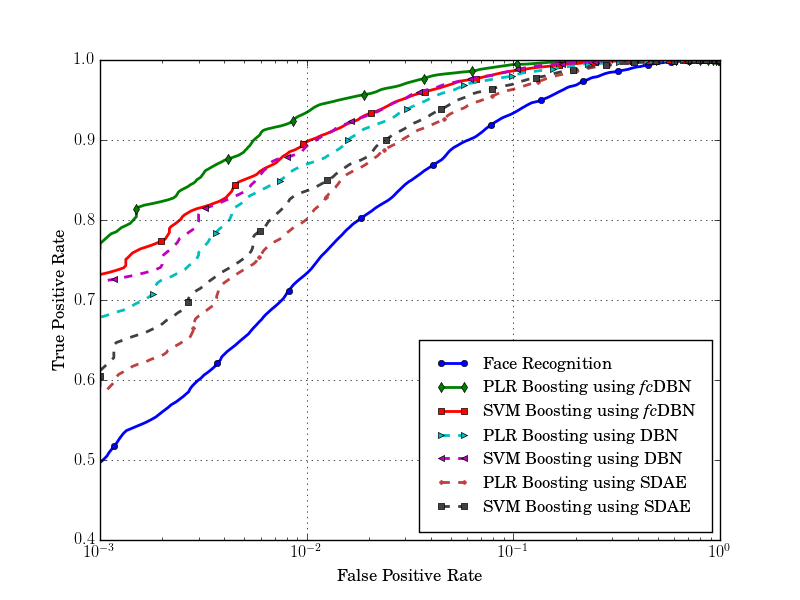}
	\label{Fig_Aided_LBP}
	%\captionsetup{justification=centering}
	\caption{ROC using LBP descriptor}
	\end{subfigure}
	\captionsetup{justification=centering}
	\caption{ROC curves summarizing the results of Kinship aided Face Verification using PLR and SVM.}
	\label{fig:Fig_AidedKinship}
\end{figure*}%

\subsection{Results of Kinship Verification}

%The results obtained with different combinations are reported in Table \ref{tab:kin_algo_table}. 
%Finally, the performance of the proposed KVRL-SDAE algorithm using the hierarchical two-stage SDAE is shown using only the top three facial regions identified in the human study. These include the complete face, T region and not-T region. 

Table \ref{tab:kin_algo_table} and Fig. \ref{fig:kin_roc} shows the results obtained using the experiments conducted on multiple databases. It is observed that KVRL-\textit{fc}DBN consistently performs better than the KVRL-SDAE and KVRL-DBN approach on all the datasets. The transformation of original input through the filters improves learning of the underlying representations. 
%\begin{table*}[!htb]
%\centering
%\small
%\begingroup
%\renewcommand*{\arraystretch}{1.2}
%\begin{tabular}{|L{3cm}|C{2cm}|C{2cm}|C{2cm}|C{2cm}|C{2cm}|C{2cm}|}
%\hline
%{\multirow{4}{*}{\textbf{Algorithm}}} & \multicolumn{6}{c|}{\begin{tabular}[c]{@{}c@{}}\textbf{Face Verification Accuracy (\%)} (TPR at 0.1\% FPR) \end{tabular}} \\ \cline{2-7} 
%                        &\multicolumn{3}{c|}{LBP Descriptor}          & \multicolumn{3}{c|}{HOG Descriptor}             \\ \cline{2-7}
%                          &  KVRL-SDAE       &  KVRL-DBN  & KVRL-\textit{fc}DBN  &  KVRL-SDAE  & KVRL-DBN  & KVRL-\textit{fc}DBN  \\ \hline
%Face Recognition   & \multicolumn{3}{c|}{49.4}          & \multicolumn{3}{c|}{59.4}                      \\ \hline
%Face Recognition Boosting with Kinship Score using PLR          & 59.3      & 67.7               & 76.4                          & 73.2     &       74.4 & 79.3                                          \\ \hline
%Face Recognition Boosting with Kinship Score using SVM          & 61.5         & 72.3       & 73.2                            & 72.9     &       76.1   & 80.0                                          \\ \hline
%\end{tabular}
%\endgroup
%\captionsetup{justification=centering}
%	\caption{Results showing improvement in face recognition accuracy when kinship score is combined with face recognition.}
%	\label{tab:Fig_AidedKinship}
%\end{table*}

Table \ref{tab:kin_table} also shows the results for different kin-relations obtained using the proposed deep learning algorithms. Compared to existing algorithms, KVRL-\textit{fc}DBN framework consistently yields state-of-the-art results and shows improvement of up to 21\% for all kin relations. It is observed that for UB database, the algorithm performs better when the images belong to children and young parents (Set 1) as compared to when there is a considerable gap between the ages of the kin (Set 2). A general trend appears for KinFace-I, KinFace-II and WVU Kinship database, where the images of kin of the same gender perform better than images belonging to a different gender. Specifically, Father-Son and Mother-Daughter kinship relations have a higher accuracy than Father-Daughter and Mother-Son. This relationship is also observed for the brothers and sisters as compared to Brother-Sister pair in the WVU Kinship database. 

The performance of the KVRL-\textit{fc}DBN approach is also computed with respect to the number of filters as shown in Fig. \ref{fig:Fig_Kinship_Results}(a). It is observed that the accuracy increases as the number of filters increases but no noticeable improvement is observed after six filters. From the human study, as mentioned previously, it is observed that the full face, T and Not-T regions are more discriminatory and thus are utilized in the KVRL-\textit{fc}DBN framework. For validation, experiments are performed by providing different regions as input to the KVRL framework and the results are shown in Fig. \ref{fig:Fig_Kinship_Results} (b). It is observed experimentally that the combination of face, T and Not-T regions perform the best in the proposed KVRL-\textit{fc}DBN framework. This approach is also computationally less intensive than using all the regions in the framework.

We also compare the performance of neural network classifier with SVM classifier for kinship verification. Using SVM with RBF kernel, across all the databases yields slightly lower performance compared to the neural network and the difference is 0.2-0.5\%. Computationally, on a six-core Xeon Processor with 64GB RAM, the proposed framework requires 1 second for feature extraction and kinship verification.

\subsection{Results of Boosting Face Verification using Kinship as Context}

The results from boosting the face verification performance using both PLR and SVM are shown in Fig. \ref{fig:Fig_AidedKinship}. It is observed that HOG descriptor performs better than LBP for face verification on the WVU Kinship dataset. However for both HOG and LBP, the face verification accuracy increases over 20\% when kinship scores obtained using the proposed KVRL-\textit{fc}DBN framework is used to boost the face verification scores. At 0.01\% FAR, a performance of 59.4\% is observed by using HOG descriptor. This improves to 79.3\% when kinship scores are utilized using \textit{fc}DBN as context and PLR algorithm is used. Similarly, the performance improves to 80.0\% when SVM is used along with \textit{fc}DBN. The improvement is more pronounced for true positive rate (TPR) at lower values of false positive rate (FPR). It is to be noted that the proposed experiment can be performed with any face verification algorithm or feature descriptor and these results suggest that incorporating kinship as soft biometric information improves the face verification performance.

%-----------------------------------------------------------------------------------------
\section{Conclusion}

The contributions of this research are four folds: (1) evaluation of human performance in kinship verification, (2) deep learning framework using proposed filtered contractive DBN (\textit{fc}DBN) for kinship verification, (3) utilizing kinship as soft biometric information for boosting face verification performance, and (4) a new kinship verification database where each subject has multiple images, that is suitable for computation of both kinship verification and kinship-aided face verification. Kin pairs having at least one female subject are found to be easily detected as kin with the pairing of \textbf{mother-son} and \textbf{sister-sister} having the two highest significance. Further, the proposed two-stage hierarchical representation learning framework (KVRL-\textit{fc}DBN) utilizes the trained deep learning representations of faces to calculate a kinship similarity score and is shown to outperform recently reported results on multiple kinship datasets. Finally, we illustrate that kinship score can be used as a soft biometric to boost the performance of any standard face verification algorithm.
As a future research direction, we can extend the proposed algorithm to build the family tree and evaluate the performance on newer kinship databases such as Family In the Wild \cite{RobinsonSWF16}.

\section*{Acknowledgments}

The authors would like to thank the associate editor and reviewers for their valuable feedback. The authors also thank Daksha Yadav for reviewing the paper. We gratefully acknowledge the support of NVIDIA Corporation for the donation of the Tesla K40 GPU utilized for this research.

\bibliographystyle{IEEEtran}

\bibliography{template_kin}

%\begin{IEEEbiography}{N}
%\end{IEEEbiography}
%\begin{IEEEbiographynophoto}{N}

\begin{IEEEbiography}[{\includegraphics[width=1in,height=1.25in,clip,keepaspectratio]{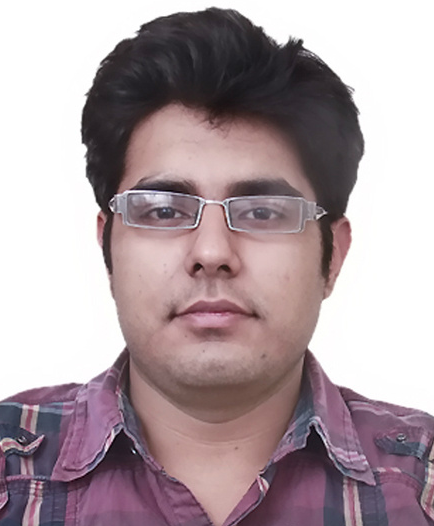}}]{Naman Kohli} received the B.Tech. (Hons.) degree in computer science from the Indraprastha Institute of Information Technology Delhi, India, in 2013. He is currently pursuing the Ph.D. degree in computer science with the Lane Department of Computer Science and Electrical Engineering at West Virginia University, USA. He is a Graduate Teaching Assistant in the Department and his research interests include biometrics, computer vision, and pattern recognition. He is a member of Phi Kappa Phi and Eta Kappa Nu honor societies. He is a recipient of the Best Paper award in IEEE Winter Conference on Applications of Computer Vision 2016.

\end{IEEEbiography}

\begin{IEEEbiography}[{\includegraphics[width=1in,height=1.25in,clip,keepaspectratio]{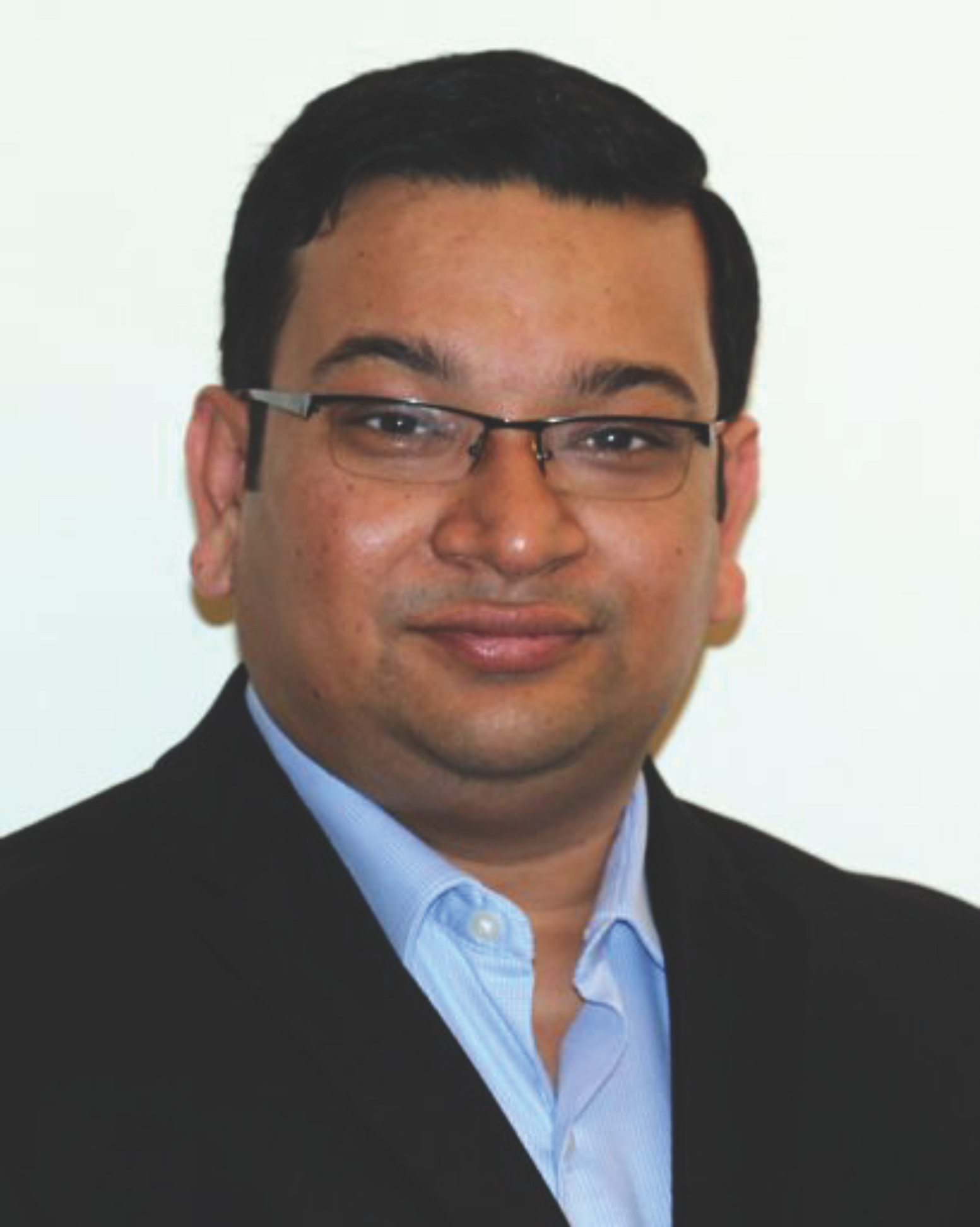}}]{Mayank Vatsa} (S'04 - M'09 - SM'14) received the M.S. and Ph.D. degrees in computer science from West Virginia University, Morgantown, USA, in 2005 and 2008, respectively. He is currently an Associate Professor with the Indraprastha Institute of Information Technology, Delhi, India and Visiting Professor at West Virginia University, USA. His research has been funded by UIDAI and DeitY, Government of India. He has authored over 175 publications in refereed journals, book chapters, and conferences. His areas of interest are biometrics, image processing, computer vision, and information fusion. He is a recipient of the AR Krishnaswamy Faculty Research Fellowship, the FAST Award by DST, India, and several best paper and best poster awards in international conferences. He is also the Vice President (Publications) of IEEE Biomertics Council, an Associate Editor of the IEEE ACCESS and an Area Editor of Information Fusion (Elsevier). He served as the PC Co-Chair of ICB 2013, IJCB 2014, and ISBA2017.

\end{IEEEbiography}

\begin{IEEEbiography}[{\includegraphics[width=1in,height=1.25in,clip,keepaspectratio]{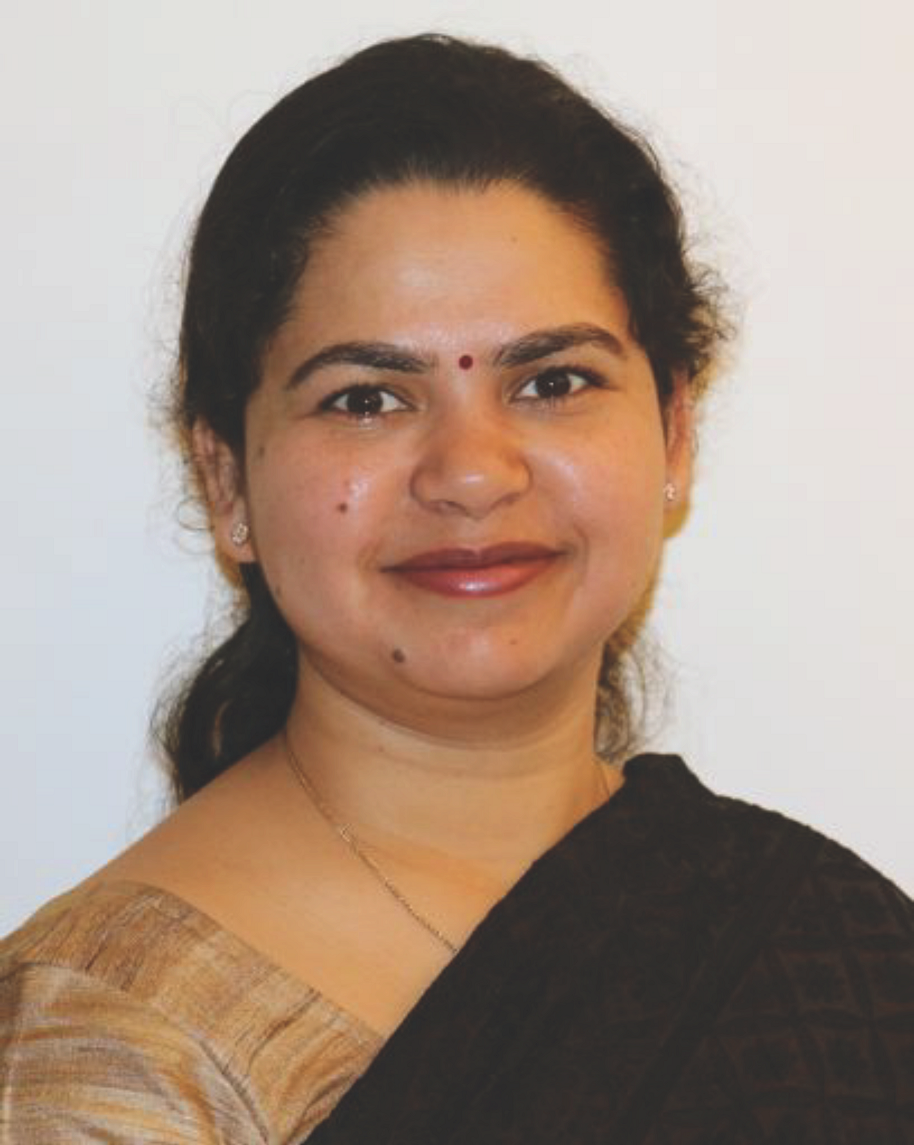}}]{Richa Singh} (S'04 - M'09 - SM'14) received the Ph.D. degree in computer science from West Virginia University, Morgantown, USA, in 2008. She is currently an Associate Professor with the Indraprastha Institute of Information Technology, Delhi, India and a Visiting Professor at West Virginia University, USA. Her research has been funded by UIDAI and MeitY, Government of India. She has authored over 175 publications in refereed journals, book chapters, and conferences. Her areas of interest are biometrics, pattern recognition, and machine learning. She is a recipient of the Kusum and Mohandas Pai Faculty Research Fellowship at the Indraprastha Institute of Information Technology, the FAST Award by DST, India, and several best paper and best poster awards in international conferences. She is also an Editorial Board Member of Information Fusion (Elsevier) and the EURASIP Journal on Image and Video Processing (Springer). She is serving as the General Co-Chair of ISBA2017 and PC Co-Chair of BTAS 2016.
\end{IEEEbiography}

\begin{IEEEbiography}[{\includegraphics[width=1in,height=1.25in,clip,keepaspectratio]{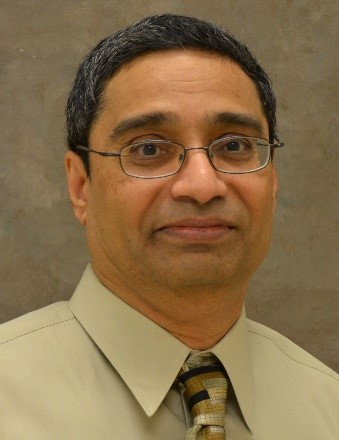}}]{Afzel Noore} received the Ph.D. degree in electrical engineering from West Virginia University. He was a Digital Design Engineer with Philips, India. From 1996 to 2003, he was the Associate Dean for Academic Affairs and a Special Assistant to the Dean with the Statler College of Engineering and Mineral Resources, West Virginia University. He is currently a Professor and the Associate Chair of the Lane Department of Computer Science and Electrical Engineering. His research interests include computational intelligence, biometrics, software reliability, machine learning, pattern recognition and image processing. His research has been funded by NASA, National Science Foundation, Westinghouse, General Electric, Electric Power Research Institute, U.S. Department of Energy, U.S. Department of Justice, and the Department of Defense Army Research Laboratory. He has authored over 120 publications in refereed journals, book chapters, and conferences. He is a member of Phi Kappa Phi, Sigma Xi, Eta Kappa Nu, and Tau Beta Pi honor societies. He is a recipient of several teaching awards and research awards at West Virginia University. He has also received eight best paper and best poster awards. He serves on the Editorial Board of the International Journal of Multimedia Intelligence and Security.
\end{IEEEbiography}

\begin{IEEEbiography}[{\includegraphics[width=1in,height=1.25in,clip,keepaspectratio]{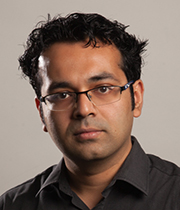}}]{Angshul Majumdar} did his Master,s and Ph.D. at the University of British Columbia in 2009 and 2012 respectively. He did his Bachelor's from Bengal Engineering College, Shibpur. Currently he is an assistant professor at Indraprastha Institute of Information Technology, Delhi. His research interests are broadly in the areas of signal processing and machine learning. He has co-authored over 120 papers in journals and reputed conferences. He is the author of Compressed Sensing for Magnetic Resonance Image Reconstruction published by Cambridge University Press and co-editor of MRI: Physics, Reconstruction and Analysis published by CRC Press. He is currently serving as the chair of the IEEE SPS Chapter's committee and the chair of the IEEE SPS Delhi Chapter.

\end{IEEEbiography}

\end{document}